\documentclass[journal]{IEEEtran}
\ifCLASSINFOpdf
\else
\fi
\usepackage{balance}
\usepackage{booktabs}
\usepackage{multirow}
\usepackage{bbding}
\usepackage{graphicx}
\usepackage{animate}
\usepackage{url}
\usepackage{color}
\usepackage{amsfonts}
\usepackage{stackrel}
\usepackage{amsmath}
\usepackage{bm}
\usepackage{arydshln}
\usepackage{tablefootnote}
\usepackage{adjustbox}
\usepackage[flushleft]{threeparttable}
\usepackage{xcolor}
\usepackage{pdfpages}

\definecolor{mygray}{RGB}{70,70,70}

\makeatletter
\renewcommand{\@IEEEsectpunct}{ \ \,}
\makeatother

\begin{document}

%

\title{PAN: Towards Fast Action Recognition via Learning Persistence of Appearance}
%
%
%

\author{Can~Zhang,
        Yuexian~Zou*,~\IEEEmembership{Senior Member,~IEEE,}
        Guang~Chen,
        and~Lei~Gan
\thanks{C. Zhang, Y. Zou, G. Chen and L. Gan are with the School of Electrical and Computer Engineering, Peking University, China (e-mail: \{zhangcan, zouyx, guangchen, ganlei\}@pku.edu.cn).}}
\maketitle

\begin{abstract}




Efficiently modeling dynamic motion information in videos is crucial for action recognition task. Most state-of-the-art methods heavily rely on dense optical flow as motion representation. Although combining optical flow with RGB frames as input can achieve excellent recognition performance, the optical flow extraction is very time-consuming. This undoubtably will count against real-time action recognition. In this paper, we shed light on fast action recognition by lifting the reliance on optical flow. Our motivation lies in the observation that small displacements of motion boundaries are the most critical ingredients for distinguishing actions, so we design a novel motion cue called Persistence of Appearance (PA). In contrast to optical flow, our PA focuses more on distilling the motion information at boundaries. Also, it is more efficient by only accumulating pixel-wise differences in feature space, instead of using exhaustive patch-wise search of all the possible motion vectors. Our PA is over 1000$\times$ faster (8196fps \emph{vs.} 8fps) than conventional optical flow in terms of motion modeling speed. To further aggregate the short-term dynamics in PA to long-term dynamics, we also devise a global temporal fusion strategy called Various-timescale Aggregation Pooling (VAP) that can adaptively model long-range temporal relationships across various timescales. We finally incorporate the proposed PA and VAP to form a unified framework called Persistent Appearance Network (PAN) with strong temporal modeling ability. Extensive experiments on six challenging action recognition benchmarks verify that our PAN outperforms recent state-of-the-art methods at low FLOPs. \emph{Codes and models are available at: \textcolor[rgb]{0,0.7,0.2}{\url{https://github.com/zhang-can/PAN-PyTorch}}}.





\end{abstract}

\begin{IEEEkeywords}
Fast Action Recognition, Motion Representation, Persistent Appearance Network, Persistence of Appearance
\end{IEEEkeywords}

%
\IEEEpeerreviewmaketitle

\section{Introduction} \label{section:intro}

\IEEEPARstart{V}{ideo} action recognition has achieved impressive performance improvements in recent years, mainly due to the aid of deep models \cite{simonyan2014two, tran2015learning,Wang_2018_CVPR} and large datasets \cite{something,kinetics,Materzynska_2019_ICCV_Workshops}. Although many prominent 2D CNNs \cite{VGG, ResNet, Inception} have been designed for image recognition task, these 2D CNNs cannot model effective dynamic motion by naively extending them to video domain. The inherent complexity of temporal evolution in videos makes \emph{motion modeling} in action recognition still a very challenging task.


\begin{figure}[t]
\begin{center}
   \includegraphics[width=1\linewidth]{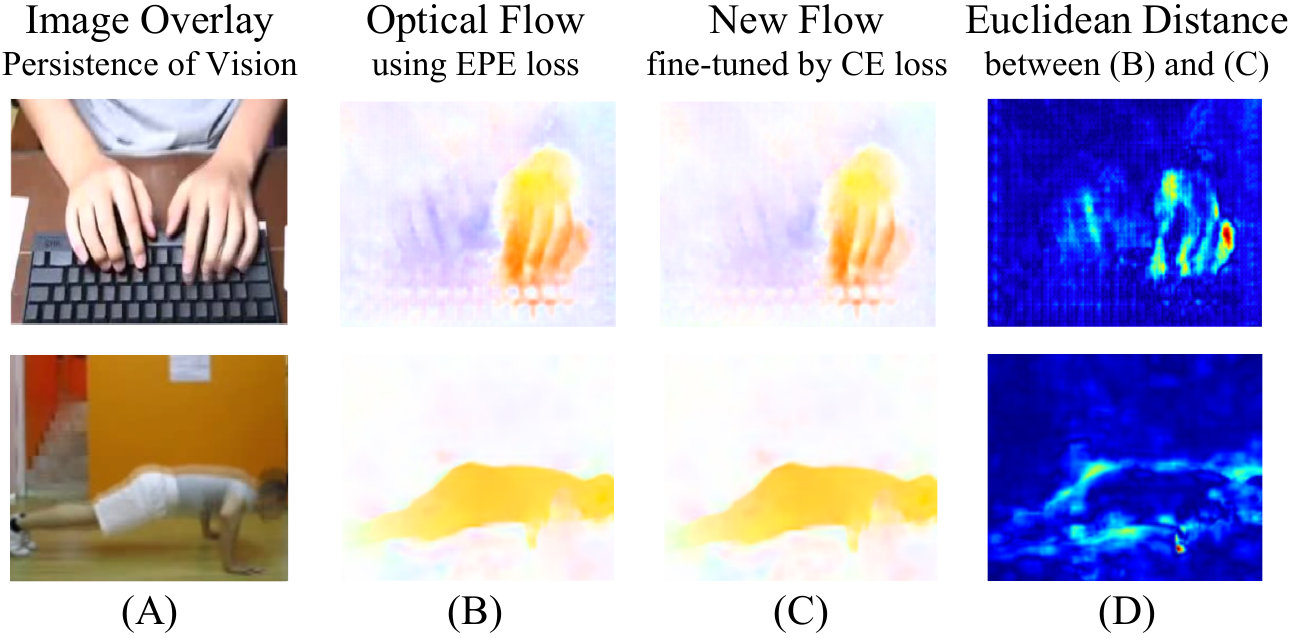}
\end{center}
   \caption{Comparison of the estimated optical flow using conventional EPE loss (B) and the new flow fine-tuned by Cross-entropy loss (C). The part (D) is obtained by calculating the Euclidean distance at each pixel between (B) and (C). In (D), the optical flow vectors change more around human motion boundaries.  \emph{Best viewed in color and zoomed in.}}
\label{fig:intro}
\end{figure}

Optical flow can encode apparent motion of moving objects in visual scenes. When combining optical flow with RGB frames as input, two-stream CNNs variants \cite{simonyan2014two,feichtenhofer2016convolutional,Zhang2018RealTimeAR,wang2016temporal,zhu2017hidden} and 3D CNNs variants \cite{ji20133d,tran2015learning,qiu2017learning,carreira2017quo} largely outperform their counterparts using only RGB frames as input. Thus, optical flow has been commonly used as motion representation for video action recognition task. However, extracting optical flow is time-costly. Aiming at mitigating this inefficiency, several recent works \cite{dosovitskiy2015flownet,ilg2017flownet,ranjan2017optical,sun2018pwc} introduce some fast and accurate optical flow estimation methods based on CNNs. But when applied to action recognition task, such solutions are still sub-optimal for two reasons: (1) Computing optical flow in advance makes action recognition a two-stage task. This two-stage paradigm is time-consuming, storage-demanding and not end-to-end trainable. (2) Improving the estimation accuracy of optical flow is not well correlated with boosting the action recognition performance, which has been demonstrated by many works \cite{sevilla2017integration,ng2018actionflownet,zhu2017hidden}. 




To address those issues, we design an alternative motion representation to replace optical flow for action recognition task. Ideally, it should be: \emph{efficient} to compute, \emph{effective} in performance, \emph{flexible} to implement and moreover, free of pre-computation and storage. To this end, we need to investigate \textbf{\emph{which parts of a moving object are most critical for distinguishing actions.}} 

Generally, most existing action recognition methods follow a two-stage procedure: they first estimate optical flow using the EPE loss\footnote{The evaluation metric of optical flow quality is end-point-error (EPE), the average Euclidean distance between the estimated and the ground-truth flow.}, and then the estimated optical flow is fed into the subsequent action recognition module. The assumption of these methods is that more accurate optical flow (measured by EPE) will bring superior action recognition performance. However, this correlation is weak as demostrated by Sevilla-Lara \emph{et al.} \cite{sevilla2017integration} and Zhu \emph{et al.} \cite{zhu2017hidden}. They combined the optical flow estimation and action recognition modules to form a unified network. The optical flow estimation module is fine-tuned by optimizing the final recognition (CE) loss instead of the EPE loss. Compared with the common two-stage methods, this unified manner achieves better recognition performance. To figure out \emph{what matters most that leads to the performance improvements}, we analyze the visualization result comparisons of the estimated optical flow using the EPE loss, the new flow fine-tuned by the CE loss and their difference computed by the Euclidean distance, as shown in Fig.~\ref{fig:intro}. The visualization results are from \cite{sevilla2017integration}. From Fig.~\ref{fig:intro}-D, we can clearly observe that the most salient parts are movement variations occurring at motion boundaries. According to the aforementioned analysis, we can conclude that \textbf{small displacements of motion boundaries} play a vital role in action recognition.


Inspired by this observation, we design a new motion cue which derives from optical flow yet focuses more on the small displacements of motion boundaries. From the perspective of human perception, a series of video frames give people a sense of motion when viewed in order at a certain speed. This phenomenon is termed as \emph{Persistence of Vision}, as shown in Fig.~\ref{fig:intro}-A blurred areas. Since we aim to extract motion information directly from RGB frames (\emph{a.k.a}, the appearance information), we name the proposed motion cue as \emph{Persistence of Appearance} (PA). Our PA enjoys high efficiency because we do not use exhaustive search of all the possible motion vectors like optical flow does. Instead, PA only contains pixel-wise operations in feature space. Specifically, given two adjacent RGB frames, our PA first computes pixel-wise intensity variations in feature space and these variations are further accumulated to manifest the motion magnitude. With such a design, PA can model the small displacements at motion boundaries because: (1) \emph{small displacements} are perceived since pixel-wise differences reflect the displacements of a small receptive field in input space; (2) \emph{motion boundaries} are captured since general patterns, \emph{e.g.}, boundaries, texture, etc, can be encoded by the first few convolutional layers \cite{zeiler2014visualizing}. So the differences among low-level feature maps can reflect the variations at boundaries.




In this way, our PA represents instantaneous motion information. However, most human actions last for a while, ranging from seconds to minutes or even longer, so long-term temporal modeling is of great importance for video action recognition. In order to aggregate the short-term dynamics contained in PA to long-term dynamics, we design a global temporal fusion strategy called \emph{Various-timescale Aggregation Pooling} (VAP). It enables the network to model long-range temporal relationships across various timescales. We further incorporate the proposed motion representation PA and the global temporal fusion strategy VAP into a unified ConvNet called \emph{Persistent Appearance Network} (PAN), which achieves fast action recognition with no upfront cost.

The preliminary work is published in ACM MM 2019 \cite{Zhang2019PANPA} and we have extended it in several significant aspects: 

\begin{itemize}
\item \emph{First}, we add more exploration studies on various network depths and investigate two encoding schemes. These new analyses further demonstrate the efficient motion modeling ability of our PA.

\item \emph{Second}, we improve the previous Various-timescale Inference Pooling (VIP) \cite{Zhang2019PANPA} to VAP. In the previous VIP, the weights of inference function for score fusion are fixed hyper-parameters. As for the enhanced version VAP, we design a more intelligent weight perception scheme that learns to adaptively aggregate the recalibrated features across different timescales. Substantial new analyses are also provided to the improved method VAP. 

\item \emph{Third}, our experiments are extended from scene-dominant datasets (Kinetics400, UCF101 and HMDB51) to more challenging temporal-dominant datasets (Something-Something-V1 \& V2 and Jester). Since temporal modeling is more critical than RGB scene information for recognizing actions in temporal-dominant datasets, the motion modeling ability of our proposed approach can be better demonstrated, \emph{i.e.}, extracting motion information directly from RGB frames.
\end{itemize}

\section{Related Work} \label{section:relatedwork}


\textbf{Network Architecture.} Spatial appearance and temporal motion are two essential ingredients for action recognition. Modeling effective spatiotemporal information still remains a challenging task. Generally, from the architectural perspective, conventional CNN-based methods can be summarized into two categories: (1) Two-stream CNNs \cite{simonyan2014two,feichtenhofer2016convolutional,Zhang2018RealTimeAR,wang2016temporal,zhu2017hidden} that separately process RGB frames (spatial stream) and pre-computed optical flow (temporal stream) using two 2D CNNs and finally apply late fusion strategy to obtain spatiotemporal semantics; (2) 3D CNNs \cite{ji20133d,tran2015learning,qiu2017learning,carreira2017quo} that jointly learn spatiotemporal features from RGB frames using 3D convolutions. However, these two paradigms are both unsatisfactory in terms of efficiency. The two-stream CNNs heavily rely on optical flow as motion representation, while the extraction of optical flow is time-consuming and storage-demanding. And the methods based on 3D CNNs is too expensive to deploy because the 3D convolution kernels require heavy computational cost.



Our PAN follows the 2D two-stream paradigm, but the input modality is only raw RGB video frames. Aiming at fast action recognition, our PAN discards the pre-computed optical flow. Instead, the motion information is distilled by introducing an efficient motion cue PA.

\textbf{Motion Representation.} Motion representation serves as an important cue for video-based action recognition. Optical flow is considered as a useful representation of short-term motion, many works \cite{simonyan2014two,wang2016temporal,ji20133d,tran2015learning,carreira2017quo} have shown that adding optical flow as another input modality can significantly boost the recognition performance. However, conventional optical flow computation approaches \cite{horn1981determining,zach2007duality,sun2010secrets} computing optical flow in advance and storing that into the disk are absolutely inefficient. In order to alleviate the inefficiency, several recent works speed up the optical flow estimation process by delicately designing some CNN models, such as FlowNet family \cite{dosovitskiy2015flownet,ilg2017flownet}, SpyNet \cite{ranjan2017optical} and PWC-Net \cite{sun2018pwc}, etc. However, these models aim at accurately estimating optical flow in advance, which is separated from the ultimate action recognition task. Other works \cite{ng2018actionflownet,zhu2017hidden} employ an encoder-decoder network to reconstruct optical flow and this network can be jointly trained with the subsequent action recognition network. But the encoder-decoder manner still requires expensive computational cost. Thus, it remains a challenging task to find an efficient and effective motion representation for action recognition. To this end, we decide to replace optical flow with an alternative motion cue for fast action recognition, rather than make optical flow estimation more accurate.


Another line of works make efforts to find auxiliary representations of motion in an end-to-end manner \cite{fan2018end, huang2018toward}. In this way, various input modalities are carefully designed, such as RGBdiff \cite{wang2016temporal}, EMV \cite{Zhang2018RealTimeAR}, dynamic image \cite{Bilen2018ActionRW} and Displacement Map \cite{zhao2018recognize}, which can be computed on-the-fly. These works still cannot perform on par with optical flow in terms of action recognition accuracy. Lee \emph{et al.} \cite{lee2018motion} proposed MFNet, which exploits five fixed directions searching strategy to encode temporal features in a unified manner.  Recently, Sun \emph{et al.} \cite{sun2018optical} introduced Optical Flow Guided Feature (OFF), which obtains spatial and temporal features utilizing Sobel operator and element-wise subtraction respectively with ground-truth optical flow as supervision. It should be noted that our work does not require optical flow during both the training and testing phases.

In this paper, we distill the motion cue PA at the bottom of the network with concise pixel-wise computation. Our PA can be viewed as feature-level pixel-wise variation accumulation, where we encode the output motion as a single channel saliency map reflecting small displacement at movement boundaries, which is of the same spatial resolution as the input RGB frames.

\textbf{Temporal Modeling.} 3D CNNs are naturally expert in temporal modeling, as 3D convolutional operators are designed to fuse both spatial and temporal information within local receptive fields. Non-local Networks \cite{Wang_2018_CVPR} use self-attention mechanism to capture long-range temporal correlations. Slowfast Networks \cite{Feichtenhofer2019SlowFastNF} contain two pathways that capture categorical semantics and motion semantics at slow and fast frame rate respectively, and lateral connections are employed to fuse the two pathways. As for 2D CNNs, 2D convolutional operators only focus on local spatial regions within single frame, without exchanging information among neighboring frames. Such frame-level features are prone to cause partial observations, thus temporal modeling is necessary. Several works \cite{feichtenhofer2016convolutional,wang2016temporal,girdhar2017actionvlad,zhou2018temporal} use 2D CNNs to process video frames independently and then obtain video-level features by late fusion strategies. Temporal Segment Networks (TSN) \cite{wang2016temporal} aggregate the frame-level features through average pooling consensus function to obtain video-level representations. But average operation can not infer the temporal order or more complicated temporal relationships. Temporal Relation Networks (TRN) \cite{zhou2018temporal} further improve the TSN to utilize temporal relations in videos. Another noteworthy work is Temporal Shift Module (TSM) \cite{lin2019tsm}, it enables local temporal fusion among neighboring frames with temporal shift operation.

In this paper, we devise a global temporal fusion strategy VAP with negligible parameters for long-term temporal modeling at multiple timescales.




\section{Persistence of Appearance (PA)} \label{section:PA}

To lift the reliance on optical flow, we devise a novel motion cue called Persistence of Appearance (PA). In this section, we first present the theoretical derivation of PA (Sec.~\ref{section:sub_PA}). Then we design an efficient PA module to speed up the motion modeling procedure (Sec.~\ref{section:PA_module}) in action recognition task. Afterwards, we investigate the function of PA in motion modeling by exploring two meaningful encoding schemes (Sec.~\ref{section:encoding}).

\subsection{Theoretical Derivation of PA} \label{section:sub_PA}

As discussed in Sec.~\ref{section:intro}, the small displacements at motion boundaries matter most for action recognition. Thus, given an adjacent two-frame pair, we make efforts to obtain a saliency map that highlights such small motion variations at boundaries. 


\begin{figure*}
\begin{center}
\includegraphics[width=1\linewidth]{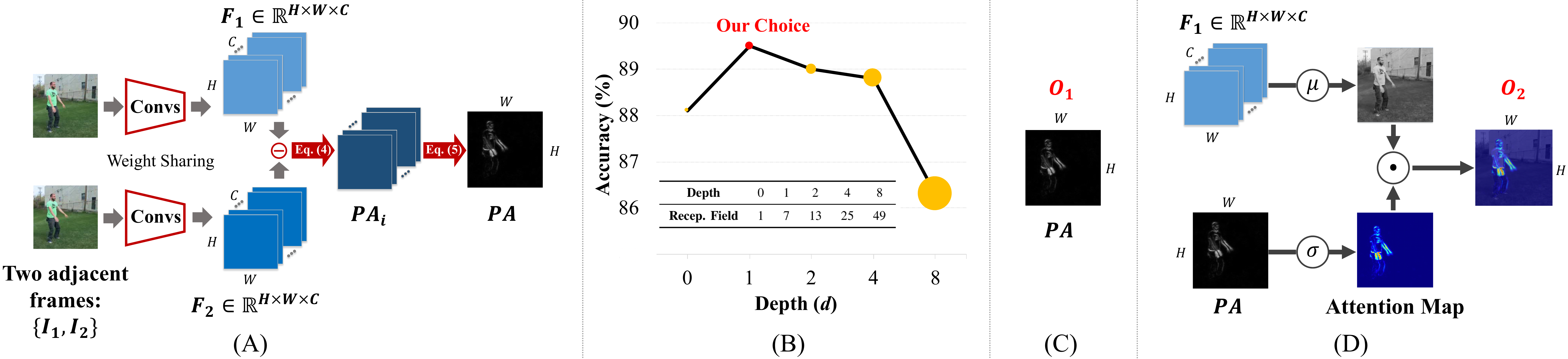}
\end{center}
   \caption{(A) Illustrations of Persistence of Appearance (PA) design. (B) Depth of conv-layers in ``PA Module'' \emph{vs.} accuracy. (C) Encoding scheme $e_1$: PA as motion modality. (D) Encoding scheme $e_2$: PA as attention. Here we only provide exemplars that processing two adjacent frames (\emph{i.e.}, $m=2$) for clarity.}
\label{fig:PA}
\end{figure*}

For traditional optical flow, the \emph{brightness constancy constraint} is defined as follows:
\begin{equation}
I(x, y, t) \approx I(x + \Delta x, y + \Delta y, t + \Delta t)
\label{equation:OF}
\end{equation}
where $I(x,y,t)$ denotes the pixel value at the location $(x,y)$ of a video frame at time $t$. As time varies from $t$ to $(t + \Delta t)$, the spatial displacements in horizontal and vertical axis are $\Delta x$ and $\Delta y$ respectively. This constraint formulation assumes that the brightness of a point remains unchanged if it moves from $(x,y)$ at time $t$ to $(x + \Delta x, y + \Delta y)$ at time $(t + \Delta t)$. The optical flow can be estimated by finding the optimal solution $(\Delta x^*, \Delta y^*)$ through optimization methods, and additional constraints, \emph{e.g.}, local smoothness assumption, are also considered for estimating the actual flow.

We extend Eq.~\ref{equation:OF} to the feature space by replacing the image $I(x,y,t)$ with its $i$-th feature map $F_i(x,y,t)$ after a specific layer:

\begin{equation}
F_i(x, y, t) \approx F_i(x + \Delta x, y + \Delta y, t + \Delta t)
\label{equation:FF}
\end{equation}

The difference map $D$ between $i$-th feature maps is given as:

\begin{equation}
D_i(x, y, \Delta t) = F_i(x + \Delta x, y + \Delta y, t + \Delta t) - F_i(x, y, t)
\label{equation:Sub}
\end{equation}

If we apply the optical flow constraint in feature space, $D$ tends to have lower absolute value. However, searching the neighboring areas to find the optimal solution $(\Delta x^*, \Delta y^*)$ in each location is time-consuming, so we do not use such a complex searching strategy. In contrast to optical flow, we only capture the motion variation at a certain point in feature space without considering the direction of the movement, which perfectly aligns with our idea of modeling the small displacements at motion boundaries because: (1) \emph{small displacements} are perceived since one pixel in low-level feature map contains information of a small receptive field in input space; (2) \emph{motion boundaries} are captured since the first few convolutional layers tend to capture general patterns, \emph{e.g.}, boundaries, texture, etc \cite{zeiler2014visualizing}. So the differences among low-level feature maps will pay more attention to the variations at boundaries. In summary, differences in low-level feature maps can reflect small displacements of motion boundaries due to convolutional operations.

Therefore, we define the $i$-th PA component as follows:
\begin{equation}
PA_{i}(p, \Delta t) = F_i(p, t + \Delta t) - F_i(p, t)
\label{equation:PA_i}
\end{equation}
where $p=(x,y)$ and $i=1,\ldots,C$, and $C$ is the channel number. Thus, we can conclude that our PA is highly correlated with optical flow. This definitely provides theoretical support for its effectiveness in modeling motion information.

All the computed PA$_i$ can be further accumulated to 1 channel to manifest the motion magnitude, which can reflect the motion variations at boundaries.

\begin{equation}
PA = \sqrt{\sum_{i=1}^C (PA_{i}(p, \Delta t))^2}
\label{equation:PA}
\end{equation}

\subsection{PA Module Design} \label{section:PA_module}

Since our PA operates in feature space, we need to search for the best depth choice of convolutional layers (conv-layers) to generate feature maps. We define the basic conv-layer as eight 7$\times$7 convolutions with stride=1 and padding=3, so that the spatial resolutions of the obtained feature maps are not reduced. Assume that $d$ basic conv-layers are sequentially stacked to form the $d$-depth network, we experiment with 5 networks having depth of $d$ equal to 0, 1, 2, 4 and 8. The experimental results on UCF101 split 1 dataset are depicted in Fig.~\ref{fig:PA}-B. The area of the circles indicates the computational cost (FLOPs). We find that directly applying the pixel-wise differences accumulation in input space ($d=0$) does not perform best. The best performance is achieved when $d=1$, \emph{i.e.}, only one basic conv-layer is adopted. As the network goes deeper, FLOPs significantly increases and the performance degrades. This is mainly because high-level features with large receptive fields have been highly abstracted and thus may not be able to reflect small motion variations in input images. The experimental results are consistent with our claim that differences in low-level feature maps can reflect small displacements of motion boundaries, which are the most critical ingredients for recognizing actions.

\begin{figure*}[!htbp]
\begin{center}
\includegraphics[width=1\linewidth]{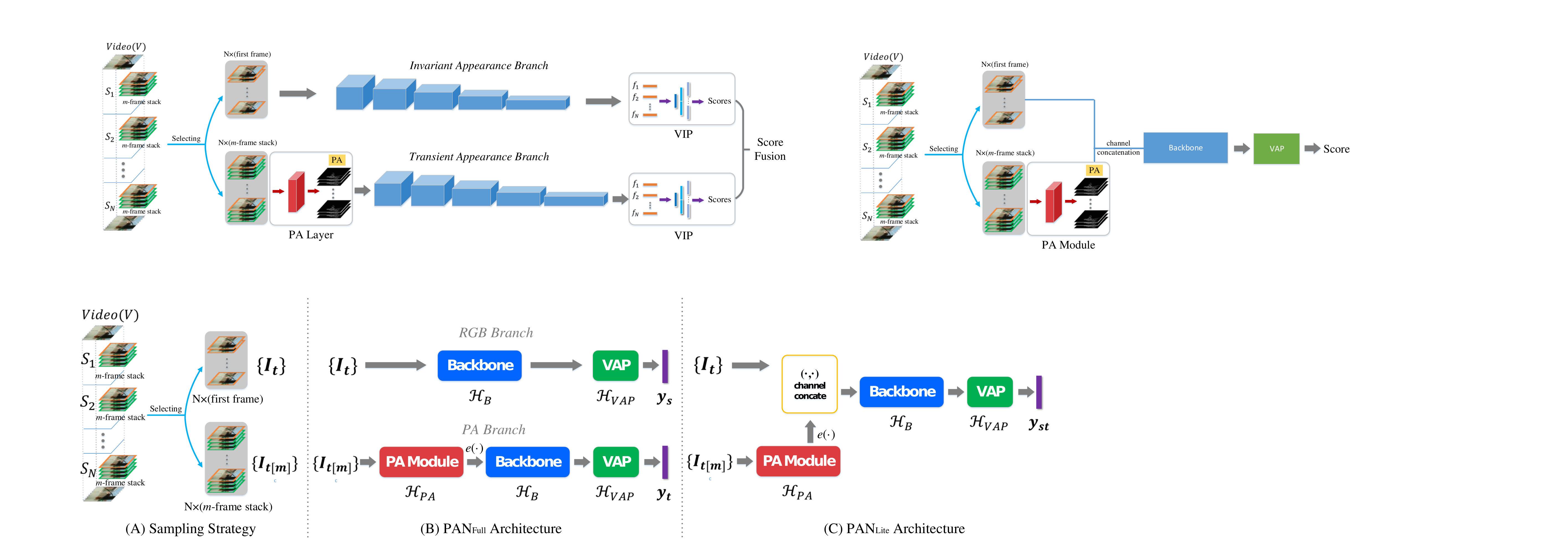}
\end{center}
   \caption{The overall architecture of Persistent Appearance Network (PAN). It has two network variants: (B) PAN$\rm _{Full}$ - ``\emph{divide and conquer}'', \emph{i.e.}, capturing the spatial and temporal semantics separately; (C) PAN$\rm _{Lite}$ - ``\emph{unified and efficient}'', \emph{i.e.}, extracting the spatial and temporal semantics simultaneously.}
\label{fig:PAN}
\end{figure*}

As $d=1$ performs best, we design a light-weighted ``PA module'' which only contains single basic conv-layer (eight 7$\times$7 convolutions) to obtain low-level features and several computing operations based on Eq.~\ref{equation:PA_i} and Eq.~\ref{equation:PA}. This module performs low-level representations comparison pixel by pixel between two adjacent frames, and outputs one saliency map (PA) reflecting small displacements of motion boundaries for further processing. This module is located at the bottom of our network, as shown in Fig.~\ref{fig:PAN}-B\&\ref{fig:PAN}-C (detailed architectural information will be given in Sec.~\ref{section:pan_full_and_lite}). 

Formally, as shown in Fig.~\ref{fig:PA}-A, given two adjacent frames $\in \mathbb{R}^{H \times W \times 3}$ with $H$, $W$ and $3$ being their height, width and channel number. First, low-level feature maps $F_1, F_2 \in \mathbb{R}^{H \times W \times C}$ are obtained without spatial resolution reduction. Then, the pixel-wise value difference is computed between the two feature maps with the same index $i$ (See in Eq.~\ref{equation:PA_i}). Finally, all the computed PA$_i$ are accumulated to 1 channel based on Eq.~\ref{equation:PA}, so the result PA $\in \mathbb{R}^{H \times W}$ is two-dimensional. Therefore, in ``PA module'', a mapping $\mathbb{R}^{H \times W \times 3} \to \mathbb{R}^{H \times W}$ is established from the appearance to the dynamic motion.


\subsection{Encoding Schemes} \label{section:encoding}


As elaborated above, PA is a concise motion cue focusing on the small displacements of motion boundaries between two adjacent frames. We would also like to understand the practical function of PA in motion modeling. Intuitively, PA can serve as either auxiliary input modality or spatial attention map. So here we explore two meaningful encoding schemes: PA as motion modality \emph{vs.} attention map. Given $m$ adjacent frames set $\lbrace I^{(i)} \rbrace_{i=1}^{m}$, the corresponding low-level feature maps in PA module are defined as $\lbrace F^{(i)} \rbrace_{i=1}^{m}$, and each two adjacent frames are processed to obtain total $(m-1)$ PA: $\lbrace PA^{(i)} \rbrace_{i=1}^{m-1}$. Assuming that the input modality to the subsequent backbone network is $O$, so in this subsection, we will discuss two encoding schemes $e_1$, $e_2$ that carry out the mapping procedure $e_i: PA \to O$, \emph{i.e.}, aggregating $PA$ to $O$.

\textbf{\emph{1) PA as motion modality.}} This is the most straightforward scheme to directly exploit motion information contained in PA. Generally, for action recognition methods, taking the stacked optical flow as input to capture motion information can significantly boost the performance. Since PA also has the capability of describing the pixel-level apparent motion information between two continuous RGB frames, we use stacked PA as input modality as shown in Fig.~\ref{fig:PA}-C. This scheme can be represented as:

\begin{equation}
O_1 = e_1(PA) = \stackrel[i=1]{m-1}{\mathrm{\Upsilon}}(PA^{(i)})
\label{equation:encoding_1}
\end{equation}

Here, we define $\stackrel[i=1]{m-1}{\mathrm{\Upsilon}}(\cdot)$ as the \emph{cumulative channel concatenation function} that chronologically concatenate the input tensor along the channel dimension. Thus, if the input tensor $PA^{(i)} \in \mathbb{R}^{H \times W \times 1}$, then the output tensor $O_1 \in \mathbb{R}^{H \times W \times (m-1)}$.

\textbf{\emph{2) PA as attention.}} Human perception researches \cite{howard2010unexpected,itti2006bayesian} suggest that instantaneous motion can attract attention. Recent video analysis works benefit a lot under the attention guidance of motion captured by optical flow, such as video salient object detection \cite{li2019motion}, video captioning \cite{chen2019motion}, etc. Motivated by this, we attempt to exploit motion information in PA to emphasize some important regions in appearance feature maps, as shown in Fig.~\ref{fig:PA}-D. This PA-guided spatial attention scheme is defined as follows, we employ PA with a sigmoid activation to attend the corresponding mean feature map:

\begin{equation}
O_2 = e_2(PA) = \stackrel[i=1]{m-1}{\mathrm{\Upsilon}}(\sigma(PA^{(i)}) \odot \mu(F^{(i)}))
\label{equation:encoding_2}
\end{equation}

\noindent where $\sigma(\cdot)$ is a sigmoid function and $\mu(\cdot)$ returns the mean value of the input feature maps along the channel dimension. $\odot$ denotes element-wise multiplication, so if the input tensor $PA^{(i)} \in \mathbb{R}^{H \times W \times 1}$ and $F^{(i)} \in \mathbb{R}^{H \times W \times C}$, then $\mu(F^{(i)}) \in \mathbb{R}^{H \times W \times 1}$ and $O_2 \in \mathbb{R}^{H \times W \times (m-1)}$.


\textbf{\emph{Which encoding scheme is better? }} We compare the performance of the PA module using these two encoding schemes in the aspect of their runtime efficiency and action recognition accuracy on UCF101 split 1 dataset. The results are shown in Table~\ref{table:comp_es}. To measure the efficiency, we consider computational cost (FLOPs), the number of parameters (\#Param) and inference speed (Speed) of the PA module. To evaluate the performance of these two encoding schemes on action recognition task, we follow the TSN manner: firstly frames are sampled from evenly divided video segments, then these frames are fed into the PA module and backbone CNN (ResNet-50) sequentially, finally the output activations are averaged as the final prediction scores. More implementation details are in the supplementary material.

The results in Table~\ref{table:comp_es} clearly indicate that encoding scheme $e_1$, directly exploiting the motion information contained in PA, performs better. It has not only fewer FLOPs but also higher runtime speed and superior recognition performance. The number of parameters of the two encoding schemes are the same, because the 1.184K parameters are completely from PA module, and the subsequent encoding procedure does not introduce any extra learnable parameters. The more FLOPs and lower speed of $e_2$ is mainly caused by the sigmoid function and element-wise multiplication. Notably, $e_2$ also degrades the accuracy by 1.5\%. We hypothesize that when using appearance-dominant features (\emph{i.e.}, appearance feature maps) as input, the features must secure integral regions of appearance to represent the semantic information for the video category. However, for $e_2$, attending appearance feature maps with PA will highlight the motion boundaries, leading to the imbalanced appearance responses both inside and at the boundaries of the moving objects, thus $e_2$ is limited in terms of such integrity. Encoding scheme $e_1$, on the contrary, only utilizes motion-dominant features (\emph{i.e.}, PA), so there is no need to consider the appearance integrity. 

\begin{table}[t]
\caption{PA as motion cue \emph{vs.} PA as attention. Accuracies are evaluated on UCF101 split 1 with the same network settings. }
\label{table:comp_es}
\begin{center}
\begin{tabular}{ccccc}
\toprule
\multirow{2.5}*{\textbf{Encoding Schemes}} & \multicolumn{3}{c}{\textbf{Efficiency Metrics}} & \multirow{2.5}*{\textbf{Accuracy}}\\
\cmidrule{2-4}
& \textbf{FLOPs} & \textbf{\#Param} & \textbf{Speed} & \\
\midrule
$e_1$: PA as motion cue & \textbf{2.868G} & \textbf{1.184K} & \textbf{8196fps} & \textbf{89.5\%}\\
$e_2$: PA as attention & 2.884G & 1.184K & 6752fps & 88.0\%\\
\bottomrule
\end{tabular}
\end{center}
\end{table}


Therefore, based on the above observations, we employ $e_1$ (\emph{i.e.}, directly exploit PA as input motion modality) as the default encoding scheme in our paper.

\section{Persistent Appearance Network (PAN)} \label{section:overview}


Our primary objective in this paper is to realize fast action recognition in real-time scenarios, so we propose the Persistent Appearance Network (PAN). In this section, we first give an architectural overview of our PAN framework (Sec.~\ref{section:pan_full_and_lite}). Then we introduce our VAP method, a temporal feature aggregation strategy, applied within PAN framework. It assists learning long-term video-level representations by integrating information across various timescales (Sec.~\ref{section:VAP}).

\subsection{PAN$_{Full}$ and PAN$_{Lite}$} \label{section:pan_full_and_lite}

The two network variants of our proposed Persistent Appearance Network (PAN) is shown in Fig.~\ref{fig:PAN}-B\&\ref{fig:PAN}-C, namely PAN$\rm _{Full}$ and PAN$\rm _{Lite}$ respectively. They have identical sampling strategy but model the spatiotemporal features in different ways. Their sampling strategy is shown in Fig.~\ref{fig:PAN}-A: the input video sequence $V$ is firstly divided into $N$ segments with equal length $\lbrace \bm{S_t} \rbrace_{t=1}^{N}$. And $m$ adjacent frames are randomly chosen from each segment as a ``$m$-frame stack'': $\lbrace \bm{I_{t[m]}} \rbrace_{t=1}^{N}$, the first frames of each ``$m$-frame stack'' are denoted as $\lbrace \bm{I_t} \rbrace_{t=1}^{N}$. The main difference between both variants are analyzed below.

\emph{1) \textbf{PAN$_{Full}$}: separate and accurate.} As illustrated in Fig.~\ref{fig:PAN}-B, this network is composed of dual branches: \emph{RGB branch} and \emph{PA branch}, capturing the spatial and temporal features separately. The RGB branch encodes the spatial appearance information. It takes the selected $N$ frames $\lbrace \bm{I_t} \rbrace_{t=1}^{N}$ as input and processes them to obtain frame-level features through the backbone network $\mathcal{H}_{B}$ (blue blocks). Then these obtained features are further aggregated as video-level features using VAP module $\mathcal{H}_{VAP}$. Mathematically, the output spatial features $\bm{y_s}$ of RGB branch can be written as:

\begin{equation}
\bm{y_s} = \mathcal{H}_{VAP}(\mathcal{H}_{B}(\lbrace \bm{I_t} \rbrace_{t=1}^{N}))
\label{equation:PAN_Full_S}
\end{equation}

The other PA branch distills apparent motion information purely from adjacent RGB frames (\emph{i.e.}, appearance information). Firstly, $N$ stacks of $m$ adjacent frames $\lbrace \bm{I_{t[m]}} \rbrace_{t=1}^{N}$ are transformed to motion cue PA after the PA module $\mathcal{H}_{PA}$. Then the temporal features $\bm{y_t}$ are obtained through the backbone network $\mathcal{H}_{B}$ and VAP module $\mathcal{H}_{VAP}$:

\begin{equation}
\bm{y_t} = \mathcal{H}_{VAP}(\mathcal{H}_{B}(e(\mathcal{H}_{PA}(\lbrace \bm{I_{t[m]}} \rbrace_{t=1}^{N}))))
\label{equation:PAN_Full_T}
\end{equation}

\noindent where $e(\cdot)$ is the encoding scheme described in Sec.~\ref{section:encoding}. Following the common practice \cite{wang2016temporal,zhou2018temporal,lin2019tsm}, we merge the branch-level scores to obtain final prediction of the whole video through score fusion strategy, \emph{i.e.}, calculating the weighted average scores from the two branches. 

\emph{2) \textbf{PAN$_{Lite}$}: unified and light-weighted.} As illustrated in Fig.~\ref{fig:PAN}-C, our unified network PAN$_{Lite}$ stacks RGB and PA together and feeds them through the backbone network, allowing the network to decide itself how to extract the spatiotemporal information. Given $N$ sampled $m$-frame stacks $\lbrace \bm{I_{t[m]}} \rbrace_{t=1}^{N}$ and their first frames $\lbrace \bm{I_t} \rbrace_{t=1}^{N}$, the output $\bm{y_{st}}$ can be written as:

\begin{equation}
\bm{y_{st}} = \mathcal{H}_{VAP}(\mathcal{H}_{B}(\bm{I_t}, e(\mathcal{H}_{PA}(\lbrace \bm{I_{t[m]}} \rbrace_{t=1}^{N}))))
\label{equation:PAN_Lite}
\end{equation}

\noindent where $(\cdot,\cdot)$ is the channel concatenation operation and $\mathcal{H}_{PA}$, $\mathcal{H}_{B}$ and $\mathcal{H}_{VAP}$ indicate PA module, backbone network and VAP module, respectively.




\begin{figure}[t]
\begin{center}
  \includegraphics[width=0.8\linewidth]{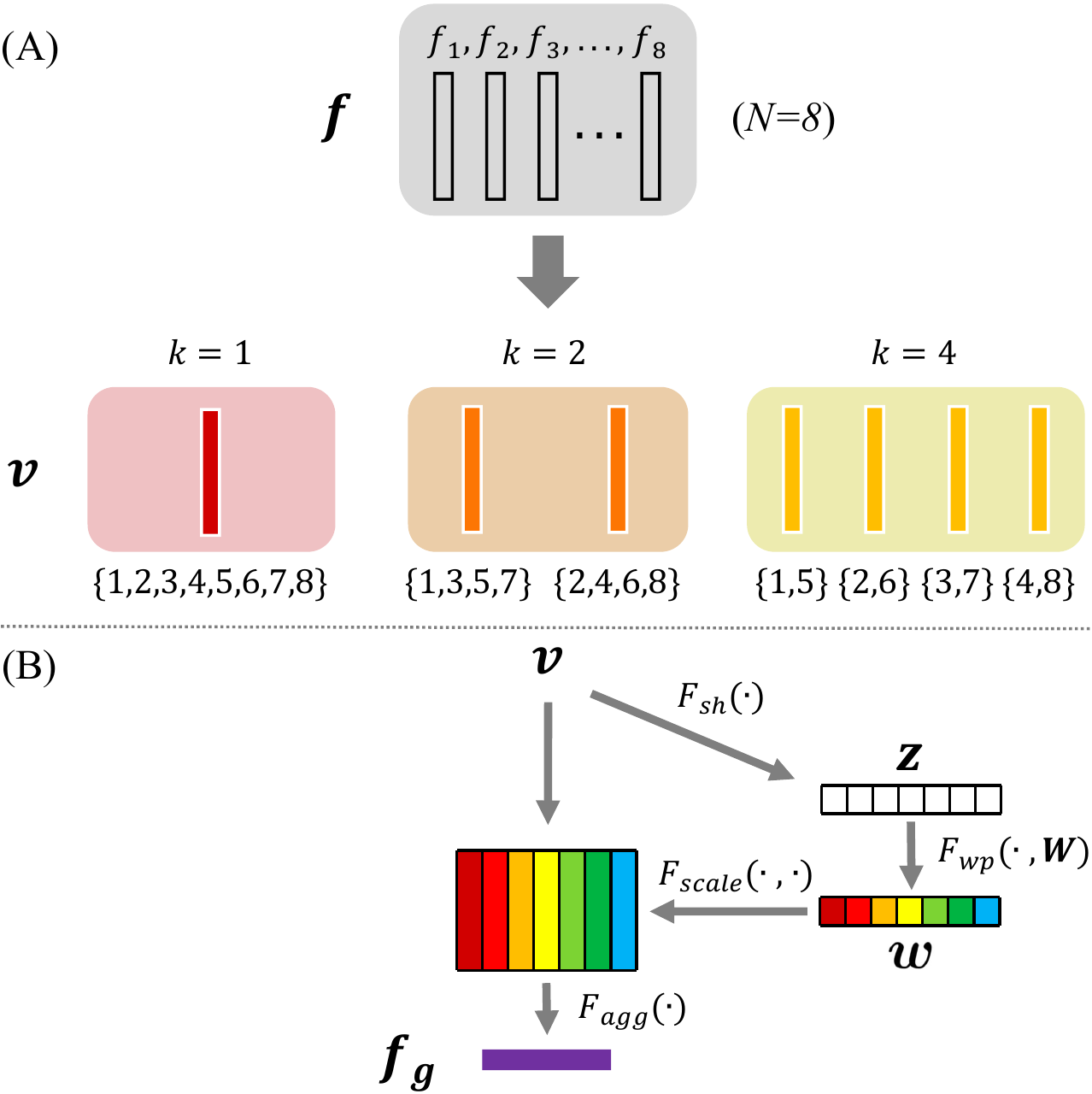}
\end{center}
   \caption{Various-timescale Aggregation Pooling (VAP). Top part: (A) Specific-timescale Pooling. Bottom part: (B) Various-timescale Aggregation. The numbers in curly brackets indicate which $\bm{f}$ are involved in generating the timescale-wise features $\bm{v}$ (Eq.~\ref{equation:pooling}). \emph{Best viewed in color and zoomed in.}}
\label{fig:VAP}
\end{figure}

\subsection{Various-timescale Aggregation Pooling (VAP)} \label{section:VAP}

Long-term temporal modeling is of great importance for the video understanding task as discussed in Sec.~\ref{section:relatedwork}. In this paper, we devise a temporal fusion strategy called Various-timescale Aggregation Pooling (VAP), which adaptively emphasizes expressive features and suppresses less informative ones by observing global information across various timescales. As shown in Fig.~\ref{fig:PAN}-B\&\ref{fig:PAN}-C, VAP module is adopted at the top of each network. Overall, VAP contains two main steps as illustrated below.

\textbf{(A) Specific-timescale Pooling}. As shown in the top part of Fig.~\ref{fig:VAP}, assume that the original video is divided into $N$ segments, and the sampled video frames are fed into the backbone 2D CNN to generate $d$-dim feature vectors $\bm{f}: \lbrace f_1, f_2, \ldots , f_N \rbrace $, where $f_i \in \mathbb{R}^{d}$. In order to temporally integrate these $N$ features without overlapped scope, we adopt dilated max pooling over the time dimension, where the dilation rate controls the spacing between the kernel points. For brevity, this pooling function can be expressed as $maxpool_{(ks, st, dr)} \lbrace activations \rbrace$, where $ks$, $st$, $dr$ refer to kernel size, stride and dilation rate respectively. Accordingly, the $k$-timescale pooling is defined as follows:
\begin{equation}
\bm{v_k} = maxpool_{(\frac{N}{k}, 1, k)}\lbrace f_1, f_2, \ldots , f_N \rbrace
\label{equation:pooling}
\end{equation}
where $k$ is a positive integer \emph{s.t.} $\frac{N}{k} \in Z$. By convention, we use pyramidal timescale settings (\emph{i.e.}, $k=2^0, 2^1, \ldots, 2^{log_{2}(N)-1}$). After Eq.\ref{equation:pooling}, the time span changes from $N$ to $k$. So $2^0+2^1+ \ldots +2^{log_{2}(N)-1} = N-1$ timescale-level features are obtained in total (\emph{i.e.}, $\bm{v} \in \mathbb{R}^{(N-1) \times d}$).



\textbf{(B) Various-timescale Aggregation}. Our basic idea is to fuse temporal information at each timescale by weighted timescale-wise aggregation. As shown in the bottom part of Fig.~\ref{fig:VAP}, given that $\bm{v} \in \mathbb{R}^{T \times d}$ represents the total $T$ pooled features at different timescales, we first \emph{shrink} global spatial semantics in each feature into a temporal descriptor reflecting the corresponding timescale-wise statistics. Thus, the output $\bm{z} \in \mathbb{R}^{T \times 1}$ can be expressed by:

\begin{equation}
\bm{z} = F_{sh}(\bm{v}) = \frac{1}{d} \sum_{i=1}^{d}\bm{v}(i)
\label{equation:squeeze}
\end{equation}

To capture the cross-timescale interdependencies, we adopt a concise nonlinearity learning mechanism with a softmax activation for \emph{weight perception}:

\begin{equation}
\bm{w} = F_{wp}(\bm{z}, \bm{W}) = softmax(\bm{W_2}(\delta(\bm{W_1z})))
\label{equation:weight_perception}
\end{equation}

\noindent where $\delta$ denotes the ReLU function, $\bm{W_1} \in \mathbb{R}^{\alpha T \times T}$ and $\bm{W_2} \in \mathbb{R}^{T \times \alpha T}$ ($\alpha$ is the expansion ratio) are the learnable parameters of two fully-connected (FC) layers. The output of $F_{wp}$ function is the weight vector $\bm{w} \in \mathbb{R}^{T \times 1}$.

The final global video-level representation $\bm{f_g} \in \mathbb{R}^{d}$ of the proposed VAP is obtained by firstly rescaling $\bm{v}$ with the weight vector $\bm{w}$ and then aggregate the recalibrated features along $T$ dimension.

\begin{equation}
\bm{f_g} = F_{agg}(F_{scale}(\bm{w}, \bm{v})) = sum(\bm{w}\bm{v})
\label{equation:aggregation}
\end{equation}

In particular, for action recognition task, the prediction scores $\bm{s}$ are obtained by:

\begin{equation}
\bm{s} = F_{pred}(\bm{f_g}, \bm{W}) = \bm{W_3} \bm{f_g}
\label{equation:prediction}
\end{equation}

\noindent where $\bm{W_3} \in \mathbb{R}^{c \times d}$, $c$ is the number of classes.

\section{Experiments}


In this section, we first introduce the evaluation datasets and implementation details. Then in ablation studies, we investigate the importance of our proposed motion cue PA and various-timescale aggregation strategy VAP for real-time action recognition. During this investigation, we also explore some basic settings. Extensive results show the superior performance achieved by PAN compared with baselines and other state-of-the-art methods, on both temporal-dominant datasets and scene-dominant datasets. Finally, we visualize the proposed motion cue PA to qualitatively justify its superiority in motion modeling and discuss the future work.

\subsection{Experimental Settings} \label{exp_set}

\textbf{Datasets.} We evaluate our approach on six challenging benchmarks for action recognition. They can be grouped into two categories: (a) \textbf{\emph{Temporal-Dominant Datasets}}, including Something-Something-V1 \& V2 \cite{something} and Jester \cite{Materzynska_2019_ICCV_Workshops}. Recognizing actions in these datasets requires strong temporal modeling ability, as many action classes are symmetrical, \emph{e.g.}, ``Moving something up'' and ``Moving something down''. Something-Something datasets (2 released versions) include 174 categories with 86,017(V1)/168,913(V2) training videos, 11,522(V1)/24,777(V2) validation videos, and 10,960(V1)/27,157(V2) test videos. Jester contains 27 human hand gestures with 148,092 videos. (b) \textbf{\emph{Scene-Dominant Datasets}}, including Kinetics400 \cite{kinetics}, UCF101 \cite{ucf101} and HMDB51 \cite{hmdb51}. RGB scene information in these datasets is more critical than temporal relations for action recognition. Kinetics400 is a large-scale action recognition benchmark including $\sim$300k videos with 400 human action classes. The performances are evaluated with the top-1 and top-5 accuracies. The UCF101 dataset includes 13,320 video clips with 101 action classes and the HMDB51 dataset contains 6,766 videos with 51 action categories. For these two datasets, the mean class accuracy over the three official splits is calculated as the final result.

\textbf{Input \& Backbone.} The input modality of our PAN is only raw RGB frames. We set $N=8$ and $m=4$ as default, thus 8 ``4-frame stack'' are sampled from a video. For PAN$\rm _{Full}$, the first frame in each ``4-frame stack'' will be selected and fed into RGB branch, meanwhile, all the frames will be fed into the PA branch. For PAN$\rm _{Lite}$, we take all the 8 sampled ``4-frame stack'' as input. For comparative purposes, we use ResNet-50 as the default backbone like other state-of-the-art methods \cite{wang2018videos,lin2019tsm,Feichtenhofer2019SlowFastNF}. Since TSM \cite{lin2019tsm} can exchange information between neighboring frames at zero cost, we opt to inject TSM into our backbone to enable \emph{local temporal fusion}. Note that TSM module is orthogonal to our PA (efficient motion representation) and VAP (\emph{global temporal fusion} strategy).


\textbf{Training \& Testing.} For relatively large-scale datasets (Something-Something-V1 \& V2, Jester and Kinetics400), the training parameters of PAN$\rm _{Lite}$ and the PA branch of PAN$\rm _{Full}$ are: initial learning rate 0.01 (total 80 epochs, divided by 10 after every 30 epochs). As for RGB branch of PAN$\rm _{Full}$, we also set initial learning rate as 0.01 (but total 50 epochs, decreases at epoch 20 \& 40). We train our PAN using SGD algorithm, with weight decay 1e-4, mini-batch size 64. ImageNet pre-trained weights are employed for initialization. While for small-scale datasets (UCF101 and HMDB51) that are easy to over-fit, we use Kinetics400 for pre-training and scale the training epochs by half. During testing procedure, we report ``view'' (spatial crops $\times$ temporal clips) used in \cite{Feichtenhofer2019SlowFastNF}. For temporal-dominant datasets, we only use 1 view (1 center spatial `224$\times$224' crop $\times$ 1 temporal clip) unless otherwise specified. While for scene-dominant datasets, we use 2 views (full resolution image with shorter side 256 pixels $\times$ 2 temporal clips), which is much more efficient than the common practice (30 views) in \cite{Wang_2018_CVPR,Feichtenhofer2019SlowFastNF,Li_2020_CVPR}. Especially, for PAN$\rm _{Full}$, we average the class prediction scores of the RGB and PA branches.


\subsection{Ablation Studies for PA} \label{section:PA_ablation}

Following the common practice \cite{wang2016temporal,feichtenhofer2016convolutional,tran2017convnet}, all experiments in this ablation study are conducted on UCF101 split 1. In this subsection, we prove that our designed PA is a significant motion representation by performing in-depth studies to answer the following two questions:

\begin{table}[t]
\caption{Comparison results of PA with mainstream optical flow computation methods. Accuracies are evaluated on UCF101 split 1 with the same network settings except the input modality. }
\label{table:comp_es_op}
\begin{center}
\begin{tabular}{ccccc}
\toprule
\multirow{2.5}*{\textbf{Motion Rep. Method}} & \multicolumn{3}{c}{\textbf{Efficiency Metrics}} & \multirow{2.5}*{\textbf{Accuracy}}\\
\cmidrule{2-4}
& \textbf{FLOPs} & \textbf{\#Param} & \textbf{Speed} & \\
\midrule


TV-L1 \cite{zach2007duality} & - & - & 8fps & 88.2\%\\
\hdashline
FlowNetS \cite{dosovitskiy2015flownet} & 356G & 38.7M & 204fps & 86.8\%\\
FlowNetC \cite{dosovitskiy2015flownet} & 444G & 39.2M & 151fps & 87.3\%\\
FlowNet2.0 \cite{ilg2017flownet} & 2019G & 162.5M & 25fps & 87.7\%\\
\midrule
\textbf{PA (Ours)} & \textbf{2.868G} & \textbf{1.184K} & \textbf{8196fps} & \textbf{89.5\%}\\
\bottomrule
\end{tabular}
\end{center}
\end{table}

\textbf{\emph{Q1: Is PA efficient, effective and flexible enough? }} As mentioned in Sec.~\ref{section:intro}, our main target in this paper is to design a motion representation alternative to optical flow with the merits of efficient, effective and flexible. We compare the PA with other mainstream optical flow extraction methods from the perspective of efficiency and effectiveness for action recognition task, including conventional optical flow \cite{zach2007duality} and CNN-based estimated optical flow \cite{dosovitskiy2015flownet,ilg2017flownet} methods. The comparison results are listed in Table~\ref{table:comp_es_op}. Note that all the network settings are the same and more details can be found in the supplementary material. Surprisingly, compared with conventional optical flow method TV-L1 \cite{zach2007duality}, our proposed PA achieves over 1000$\times$ faster speed (8196fps \emph{vs.} 8fps) as well as 1.3\% higher action recognition accuracy. Furthermore, PA consistently outperforms all the seminal CNN-based optical flow estimation methods \cite{dosovitskiy2015flownet,ilg2017flownet}, suggesting that PA has fast (8196fps) and effective (89.5\%) motion modeling ability.

To demonstrate the flexibility of our proposed PA, we deploy different backbone networks and compare their results. Here, we choose three widely adopted backbone networks for deep action recognition, including BN-Inception \cite{ioffe2015batch} (2D CNN), ECO \cite{zolfaghari2018eco} (mixed 2D-3D CNN) and 3D-ResNet \cite{tran2017convnet} (3D CNN). Experimental results on UCF101 split 1 are tabulated in Table~\ref{table:flexibility}. The results show that the selected backbone networks all significantly benefit from our PA with considerable accuracy improvements, demonstrating the flexibility of our proposed PA.

Therefore, compared with optical flow, our proposed PA is much faster to compute, superior in performance. And it is also flexible to implement.

\begin{table}[t]
\caption{Flexibility validation of PA. Our motion cue PA considerably improves all the baselines on UCF101 split 1. All the experiments are conducted using the same network settings.}
\label{table:flexibility}
\begin{center}
\begin{tabular}{ccccc}
\toprule
\textbf{Baseline} & \textbf{Pre-train} & \textbf{Modality} & \textbf{Acc.} & \textbf{$\Delta$Acc.}\\
\midrule
\multirow{2}*{BN-Inception \cite{ioffe2015batch}} & \multirow{2}*{ImageNet} & RGB & 85.6\% & \multirow{2}*{\textbf{+5.7\%}}\\
& & RGB+PA & \textbf{91.3\%} & \\
\midrule
\multirow{2}*{ECO \cite{zolfaghari2018eco}} & \multirow{2}*{Kinetics} & RGB & 90.4\% & \multirow{2}*{\textbf{+4.7\%}}\\
& & RGB+PA & \textbf{95.1\%} & \\
\midrule
\multirow{2}*{3D ResNet-18 \cite{tran2017convnet}} & \multirow{2}*{Kinetics} & RGB & 83.9\% & \multirow{2}*{\textbf{+2.9\%}}\\
& & RGB+PA & \textbf{86.8\%} & \\
\bottomrule
\end{tabular}
\end{center}
\end{table}

\begin{figure}[t]
\begin{center}
\includegraphics[width=1\linewidth]{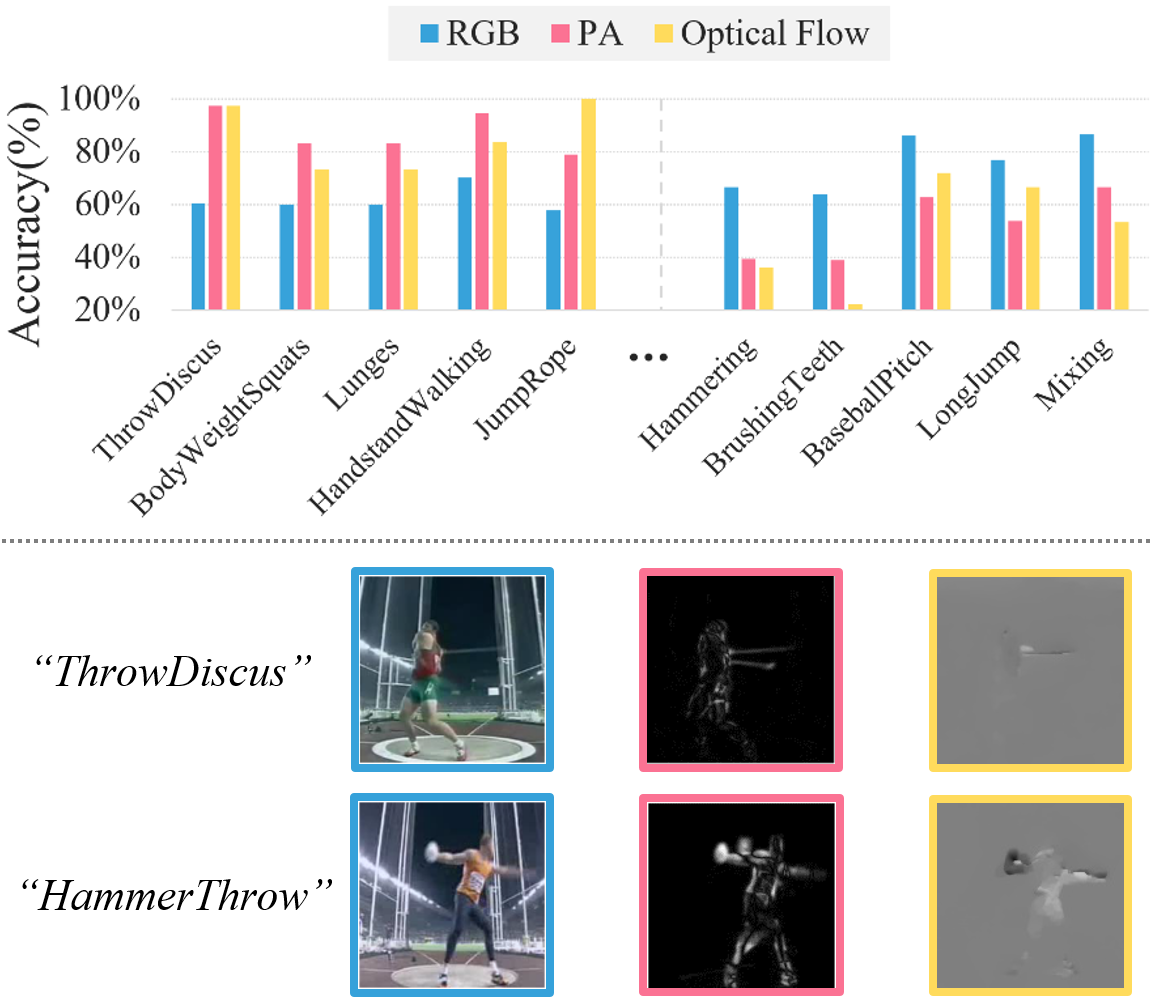}
\end{center}
   \caption{Top part: The top 5 action classes that show the greatest difference in terms of accuracy between the input modality RGB and PA, and the performance of optical flow in these classes is also plotted for comparison. Bottom part: Input modality visualization of the two classes that are most easily misclassified by RGB. \emph{Best viewed in color and zoomed in.}}
\label{fig:top5}
\end{figure}

\textbf{\emph{Q2: Can PA represents apparent motion? }} A key intuition for designing PA is that it can capture apparent motion by focusing more on moving boundaries without the heavy pre-computation of optical flow. In order to investigate whether our PA performs similarly to the optical flow, we compare the performance using three different input modalities: RGB, PA and optical flow. With each modality as input, we train the network first on UCF101 split 1, then we test the trained models to get the prediction scores for all the 101 classes. Finally, we plot the top 5 classes that show the largest differences in recognition accuracy between RGB and PA. For reference, the performance of optical flow in these classes is also depicted, as shown in the top part of Fig.~\ref{fig:top5}. In this figure, we can clearly see that when PA outperforms RGB, the optical flow is also superior to RGB and vice versa. This indicates that PA can help the network learn patterns different from RGB but similar to optical flow. 

\begin{figure*}[!htbp]
\begin{center}
   \includegraphics[width=1\linewidth]{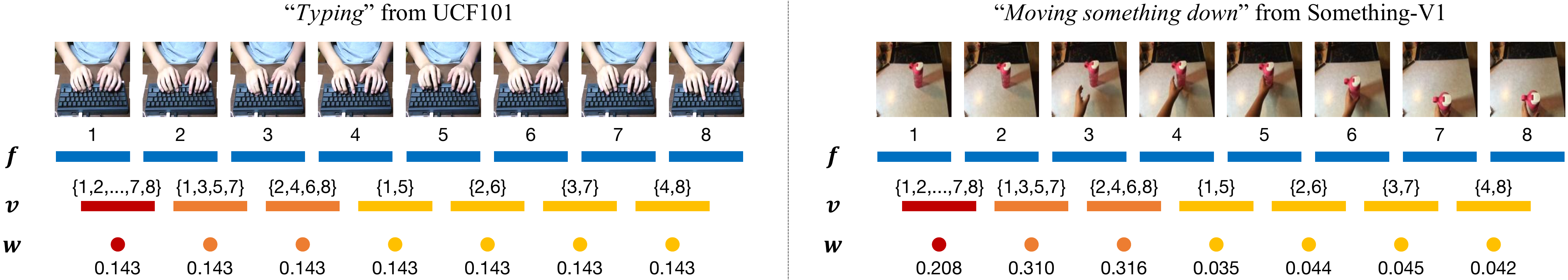}
\end{center}
   \caption{Visualization of two selected video sequences and their timescale-wise weights $\bm{w}$ calculated by Eq.~\ref{equation:weight_perception}. The sampled frames are processed by the backbone individually to obtain frame-level features $\bm{f}$. The numbers in curly brackets indicate which $\bm{f}$ are involved in generating the timescale-wise features $\bm{v}$ (Eq.~\ref{equation:pooling}). Different colors (red, orange \& yellow) are employed to distinguish different timescales. \emph{Best viewed in color and zoomed in.}}
\label{fig:VAP_ablation}
\end{figure*}

After analyzing the recognition results, we find that most ``Throw Discus'' videos are misclassified as ``Hammer Throw'' when using just RGB modality. To demonstrate that our PA can represent apparent motion, we visualize the three modalities of these two classes in the bottom part of Fig.~\ref{fig:top5}. In RGB images, it is unclear whether the athletes are holding the discus or the hammer due to the interference of background, illumination, etc, so distinguishing the two classes using only RGB modality may cause confusion. Encouragingly, our PA is able to highlight the moving objects (in this case human and the object in hands) as the optical flow does, which is crucial for differentiating between videos that look much alike. 

Therefore, this result is in accordance with our idea that the PA, focusing on capturing small displacements at moving boundaries, can be used as apparent motion representation. See Sec.~\ref{section:visualization} for more visualization results.



\begin{table}[t]
\caption{Exploration of different expansion ratio $\alpha$ and comparison among different temporal aggregation strategies.}
\label{table:AP_VIP_VAP}
\begin{center}
\begin{tabular}{cc}
\toprule
\textbf{Aggregation Strategy} & \textbf{Accuracy}\\
\midrule
Avg Pooling (baseline) & 86.5\% \\
\midrule
VIP \cite{Zhang2019PANPA} (Our prev.) & 87.9\% \\
\midrule
VAP($\alpha=1$) & 88.4\% \\
VAP($\alpha=2$) & 88.3\% \\
\textbf{VAP($\bm{\alpha=4}$)} & \textbf{88.5\%} \\
VAP($\alpha=8$) & 88.4\% \\
\bottomrule
\end{tabular}
\end{center}
\end{table}

\subsection{Ablation Studies for VAP} \label{section:VAP_ablation}

As introduced in Sec.~\ref{section:VAP}, our proposed VAP is a temporal aggregation strategy that can learn to exploit global information across different timescales. In this subsection, we conduct ablation experiments to gain more insights about the effect of aggregating semantics across various timescales.

\textbf{\emph{Q1: Is various-timescale aggregation effective? }} Following the common practice \cite{wang2016temporal,feichtenhofer2016convolutional,tran2017convnet}, we perform ablation studies on UCF101 split 1 dataset using different temporal aggregation strategies. $N=8$ frames are sampled as input and ResNet-50 is used as backbone network. The results are shown in Table~\ref{table:AP_VIP_VAP}. Avg Pooling here is referred to the conventional consensus function introduced in \cite{wang2016temporal} that averages the output logits to get the final prediction. First, we explore the impact of the expansion ratio $\alpha \in \lbrace 1,2,4,8\rbrace$. The accuracies between various expansion ratios exhibit stable performances, consistently outperforming the Avg Pooling baseline. This indicates that VAP is not sensitive to the hyperparameter setting $\alpha$. As $\alpha=4$ achieves the best performance, we choose 4 as the default expansion ratio in VAP. Second, VAP is considered as an enhanced version of our previous work VIP \cite{Zhang2019PANPA}. Different from VIP that exploits preset weights for inference, VAP learns the weights by adaptively using global information at different timescales. We also do comparison between these two variants. VAP achieves higher recognition accuracy than VIP (88.5\% \emph{vs.} 87.9\%), showing the effectiveness of the enhancement in this paper. Furthermore, VAP and VIP utilizing various-timescale semantics consistently outperform baseline by 2.0\% and 1.4\% respectively, indicating the significance of long-term temporal modeling through various-timescale aggregation.



\begin{table*}[htbp]
\caption{Comparison results of PAN with other state-of-the-art methods on Something-Something V1 \& V2 and Jester datasets.}
\label{table:sthv1_comp}
\begin{center}
\begin{adjustbox}{center}
\begin{threeparttable}
\begin{tabular}{ccccccccccc}
\toprule
\multirow{3.5}*{\textbf{Methods}} & \multirow{3.5}*{\textbf{Backbone}} & \multirow{3.5}*{\textbf{Flow?}} & \multirow{3.5}*{\textbf{\#Frame}} & \multirow{3.5}*{\textbf{FLOPs$\times$views}} & \multicolumn{2}{c}{\textbf{Something-V1}} & \multicolumn{2}{c}{\textbf{Something-V2}}& \multicolumn{2}{c}{\textbf{Jester}}\\
\cmidrule(lr){6-7}
\cmidrule(lr){8-9}
\cmidrule(lr){10-11}
& & & & &\textbf{Val} & \textbf{Val} & \textbf{Val} & \textbf{Val} & \textbf{Val} & \textbf{Val}\\
& & & & &\textbf{Top1} & \textbf{Top5} & \textbf{Top1} & \textbf{Top5} & \textbf{Top1} & \textbf{Top5} \\
\midrule

I3D \cite{wang2018videos} & \multirow{2}*{3DResNet-50} & & \multirow{3}*{32$\times$2} & 153G$\times$2 & 41.6 & 72.2 & - & - & - & -\\
NL I3D \cite{wang2018videos} & & & & 168G$\times$2 & 44.4 & 76.0 & - & - & - & -\\
NL I3D + GCN \cite{wang2018videos} & 3DResNet-50+GCN & & & 303G$\times$2 & 46.1 & 76.8 & - & - & - & -\\
\hdashline
ECO$\rm _{En}$Lite \cite{zolfaghari2018eco} & BNInception & & 92 & 267G$\times$1 & 46.4 & - & - & - & - & -\\
{\color{mygray}ECO$\rm _{En}$Lite$\rm _{RGB+Flow}$ \cite{zolfaghari2018eco}} & +3DResNet-18 & \Checkmark & {\color{mygray}92+92$\times$6} \textcolor{red}{\textsuperscript{$\dagger$}} & {\color{mygray}N/A} & {\color{mygray}49.5} & {\color{mygray}-} & {\color{mygray}-} & {\color{mygray}-} & {\color{mygray}-} & {\color{mygray}-}\\
\midrule

\multirow{2}*{TSN$\rm _{8F}$ \cite{wang2016temporal}} & BNInception & & 8 & 16G$\times$1 & 19.5 & - & - & - & - & -\\
& ResNet-50 & & 8 & 33G$\times$1 & 19.7 & 46.6 & 27.8 & - & 81.0 & 99.0 \\
\hdashline
\multirow{2}*{TRN-Multiscale \cite{zhou2018temporal}} & BNInception & & 8 & 16G$\times$1 & 34.4 & - & 48.8 & 77.6 & 95.3 & -\\
& ResNet-50 & & 8 & 33G$\times$1 & 38.9 & 68.1 & - & - & - & -\\
{\color{mygray}TRN$\rm _{RGB+Flow}$ \cite{zhou2018temporal}} & {\color{mygray}BNInception} & \Checkmark & {\color{mygray}8+8$\times$6} \textcolor{red}{$\dagger$} & {\color{mygray}N/A} & {\color{mygray}42.0} & {\color{mygray}-} & {\color{mygray}55.5} & {\color{mygray}83.1} & {\color{mygray}-} & {\color{mygray}-}\\
\hdashline
TSM$\rm _{8F}$ \cite{lin2019tsm} & \multirow{3}*{ResNet-50} & & 8 & 33G$\times$1 & 45.6 & 74.2 & 59.1 & - & 94.4 & 99.7\\
TSM$\rm _{16F}$ \cite{lin2019tsm} & & & 16 & 65G$\times$1 & 47.2 & 77.1 & 63.4 & 88.5 & 95.3 & 99.8\\
{\color{mygray}TSM$\rm _{RGB+Flow}$ \cite{lin2019tsm}} & & \Checkmark & {\color{mygray}16+16$\times$6} \textcolor{red}{\textsuperscript{$\dagger$}} & {\color{mygray}N/A} & {\color{mygray}52.6} & {\color{mygray}81.9} & {\color{mygray}66.0} & {\color{mygray}90.5} & {\color{mygray}-} & {\color{mygray}-}\\
\hdashline
TEA$\rm _{8F}$ \cite{Li_2020_CVPR} & \multirow{2}*{ResNet-50} & & 8 & 35G$\times$30 & 51.7 & 80.5 & - & - & - & -\\
TEA$\rm _{16F}$ \cite{Li_2020_CVPR} & & & 16 & 70G$\times$30 & 52.3 & 81.9 & - & - & - & -\\
\midrule

PAN$\rm _{Lite}$ (Ours) & \multirow{3}*{ResNet-50+TSM} & & 8+8$\times$4 \textcolor{red}{$\dagger$} & 35.7G$\times$1 & 48.0 & 76.1& 60.8 & 86.7 & 96.2 & 99.8\\
PAN$\rm _{Full}$ (Ours) & & & 8+8$\times$4 \textcolor{red}{\textsuperscript{$\dagger$}} & 67.7G$\times$1 & 50.5 & 79.2 & 63.8 & 88.6 & 96.6 & 99.8\\
PAN$\rm _{En}$ (Ours) & & & (8+8$\times$4)$\times$2 & (46.6G+88.4G)$\times$2 & 53.4 & 81.1 & 66.2 & 90.1 & 97.2 & 99.9\\
\hdashline
PAN$\rm _{En}$ (Ours) & ResNet-101+TSM & & (8+8$\times$4)$\times$2 & (85.6G+166.1G)$\times$2 & \textbf{55.3} & \textbf{82.8} & \textbf{66.5} & \textbf{90.6} & \textbf{97.4} & \textbf{99.9}\\
\bottomrule
\end{tabular}
\begin{tablenotes}
\footnotesize
\item \textcolor{red}{$\dagger$} The flow streams of \cite{zolfaghari2018eco,zhou2018temporal,lin2019tsm} take 10-channel (from ``6-frame stack'') optical flow as input, while our PAN only uses ``4-frame stack''.
\end{tablenotes}
\end{threeparttable}
\end{adjustbox}
\end{center}
\end{table*}

\textbf{\emph{Q2: What does the VAP learn? }} We expect VAP module to let the network capture various-timescale interdependencies. To verify that this is indeed achieved, we output the weight results $\bm{w}$ learned from the weight perception Eq.~\ref{equation:weight_perception}. Here we consider both scene-dominant dataset (UCF101) and temporal-dominant dataset (Something-Something-V1). $N=8$ frames are sampled as input, thus $k \in \lbrace 1,2,3 \rbrace$ and total 7 timescale-level features $\bm{v}$ are calculated according to Eq.~\ref{equation:pooling}. ResNet-50 is used as backbone and VAP module is adopted at the top of the network. We first train this network on the two datasets respectively, then we test the two trained models separately to obtain the corresponding weight values $\bm{w}$ of different timescales. Fig.~\ref{fig:VAP_ablation} shows two selected video sequences and the timescale-wise weights $\bm{w}$ with respect to the two action classes: ``\emph{Typing}'' from UCF101 and ``\emph{Moving something down}'' from Something-Something-V1 dataset. More visualization results can be found in the supplementary material. We can observe that the weight changes for different videos at various timescales, suggesting that VAP can adaptively guide the network to emphasize some expressive features while suppressing other less informative ones. Moreover, we can clearly see that VAP trained on UCF101 dataset tends to learn similar weights for all timescales, while the weights varies widely on Something-Something-V1. We speculate that this has to do with the fact that temporal relationships in Something-Something-V1 is more crucial than that in UCF101 for recognition \cite{zhou2018temporal,lin2019tsm,Xie2018RethinkingSF}, so VAP module trained on Something-Something-V1 prefers treating all timescales differently. This observation further proves that various-timescale aggregation is an important ingredient for video temporal semantics modeling.



\subsection{Comparison with the State-of-the-Arts}

To validate the efficacy of our approach, we compare PAN with newly published state-of-the-art methods on six datasets. Since our proposed PAN focuses on temporal semantics modeling, we mainly analyze the results on temporal-dominant datasets (Something-Something-V1 \& V2 and Jester). We also evaluate PAN on various scene-dominant datasets (Kinetics400, UCF101 and HMDB51) to show its consistent strong performance.

\emph{\textbf{1) Temporal-Dominant Datasets.}}

\textbf{Something-Something-V1 \& V2 and Jester}. Table~\ref{table:sthv1_comp} shows the comparison results with the state-of-the-arts on Something-Something-V1 \& V2 and Jester datasets. We group the listed methods into two sections (separated by solid line) according to their backbone types: 3D CNN based methods \cite{wang2018videos,zolfaghari2018eco} and 2D CNN based methods \cite{wang2016temporal,zhou2018temporal,lin2019tsm}. It is clear that with complex 3D convolutional operations, the FLOPs of I3D \cite{wang2018videos} and ECO \cite{zolfaghari2018eco} are much higher than the 2D CNN based methods. Our PAN surpasses all the 3D CNN based methods in both efficiency and accuracy aspects. For example, compared with NL I3D+GCN \cite{wang2018videos}, our PAN$\rm _{Lite}$ achieves 1.9\% higher top-1 accuracy (48.0\% \emph{vs.} 46.1\% on Something-V1) while with only $\sim$6\% computational cost (35.7G \emph{vs.} 303G$\times$2). As for 2D CNN based methods, the performance of the baseline TSN \cite{wang2016temporal} is relatively inferior than other methods, indicating the significance of temporal modeling for these datasets. Compared with the efficient (33G FLOPs) baselines TSN$\rm _{8F}$ \cite{wang2016temporal} and TSM$\rm _{8F}$ \cite{lin2019tsm}, our PAN$\rm _{Lite}$ achieves much higher top-1 accuracy (48.0\% \emph{vs.} 19.7\%/45.6\% on Something-V1, 60.8\% \emph{vs.} 27.8\%/59.1\% on Something-V2, 96.2\% \emph{vs.} 81.0\%/94.4\% on Jester) with only slight extra cost (0.08$\times$ FLOPs). With dual-path structure that separately processes RGB and PA modalities, PAN$\rm _{Full}$ further bolsters the action recognition performance (50.5\% on Something-V1, 63.8\% on Something-V2 and 96.6\% on Jester). The consistent improvements of our PAN over the other state-of-the-art methods strongly justify the superiority of our proposed PA and VAP for temporal modeling. 

Following \cite{zolfaghari2018eco,lin2019tsm}, we also average the class prediction results of PAN$\rm _{Lite}$ and PAN$\rm _{Full}$ to form our ensemble version PAN$\rm _{En}$. Here, two views (1 full-resolution with 256 shorter size $\times$ 2 temporal clips) are used for inference. Compared with the optical flow based methods TRN$\rm _{RGB+Flow}$ (42.0\%), ECO$\rm _{En}$Lite$\rm _{RGB+Flow}$ (49.5\%) and TSM$\rm _{RGB+Flow}$ (52.6\%), our PAN$\rm _{En}$ (53.4\%) provides +11.4\%, +3.9\% and +0.8\% top-1 accuracy improvements on Something-V1, respectively. As listed in Table~\ref{table:comp_es_op}, the mainstream optical flow extraction methods are computationally intensive, even the most light-weighted FlowNetS model needs 356G FLOPs and extra storage. Taking this cost into account, the optical flow based methods are expensive in both time and space. Surprisingly, the total cost of our PAN$\rm _{En}$ model is only $\sim$270G. This observation confirms that our proposed PAN, with an efficient motion cue PA, can effectively accelerate the video representation learning process by lifting the reliance on optical flow.

Furthermore, when exploiting a deeper backbone ResNet-101, our PAN brings the current state-of-the-art results to a whole new level (55.3\% on Something-V1, 66.5\% on Something-V2, 97.4\% on Jester). Notably, our method only takes RGB frames as input and thus is free of optical-flow pre-computation.



\begin{figure*}[!htbp]
\begin{center}
   \includegraphics[width=1\linewidth]{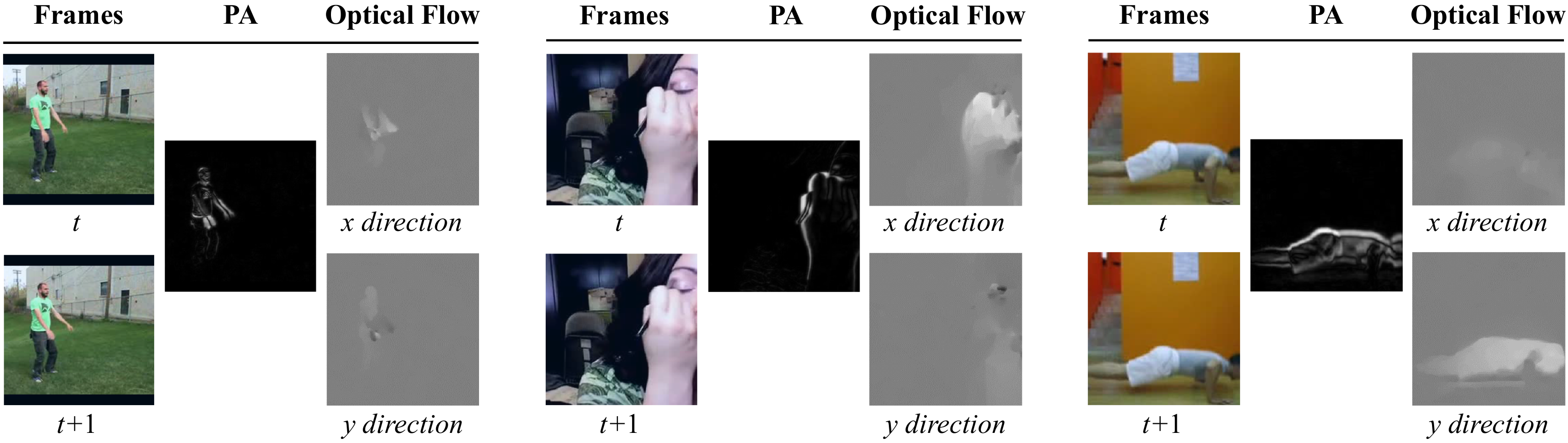}
\end{center}
   \caption{Visualization of two adjacent frames and their corresponding PA and optical flow. Left: \emph{BodyWeightSquats}. Middle: \emph{ApplyEyeMakeup}. Right: \emph{PushUps}. \emph{Best viewed in color and zoomed in.}}
\label{fig:vis}
\end{figure*}

\emph{\textbf{2) Scene-Dominant Datasets.}}

\textbf{Kinetics400, UCF101 and HMDB51}. We also compare the performance of our PAN with other recent state-of-the-art methods on three scene-dominant datasets: Kinetics400, UCF101 and HMDB51. The results are summarized in Table~\ref{table:scene_d_comp}. Our method achieves very competitive performance on these datasets, where most actions can be classified by a single frame (\emph{e.g.}, ``Typing'' action shown in Fig.~\ref{fig:VAP_ablation}). Our PAN outperforms most of the heavy 3D CNN based architectures (first part in Table~\ref{table:scene_d_comp}) \cite{Diba2018SpatioTemporalCC,zolfaghari2018eco,carreira2017quo} using 2D CNN as backbone, indicating that it can enhance the traditional 2D CNN with low cost. For 2D CNN based methods (second part), when pursuing high accuracy, they usually sample many views (spatial crops $\times$ temporal clips) per video for inference, thus leading to expensive computational cost. For example, TSM \cite{lin2019tsm} and TEA \cite{Li_2020_CVPR} sample 30 views, their FLOPs are even higher than that of 3D CNN based architectures. In contrast, we only use 2 views for efficiency concerns (detailed in Sec.~\ref{exp_set}) and achieve superior performance. For example, compared with baseline TSM \cite{lin2019tsm}, our PAN$\rm _{En}$ achieves +1.2\%, +1.3\% and +3.8\% top-1 accuracy improvements on Kinetics400, UCF101 and HMDB51 respectively with only $\sim$20\% computational cost (270G \emph{vs.} 1290G). Notably, our PAN is even superior to the optical flow based methods (third part) except for I3D \cite{carreira2017quo}. Since I3D is based on heavy 3D convolutions and takes pre-computed optical flow as input, its computational cost is much more expensive than our PAN.

\begin{table}[t]
\caption{Comparison results of PAN with other state-of-the-art methods on Kinetics400, UCF101 and HMDB51 datasets. }
\label{table:scene_d_comp}
\begin{center}
\begin{adjustbox}{center}
\begin{threeparttable}
\begin{tabular}{ccccc}
\toprule
\multirow{2}*{\textbf{Method}} & \textbf{FLOPs}
& \textbf{Kinetics400} & \multirow{2}*{\textbf{UCF101}} & \multirow{2}*{\textbf{HMDB51}}\\
& \textbf{$\times$views} & \textbf{top1 (top5)} & & \\
\midrule
STC \cite{Diba2018SpatioTemporalCC} & - & 68.7 (88.5) & 93.7 & 66.8\\
ECO$\rm _{En}$ \cite{zolfaghari2018eco} & 368G$\times$1 & 70.0 (89.4) & 94.8 & 72.4\\
I3D$\rm _{RGB}$ \cite{carreira2017quo} & 108G$\times$- & 71.1 (89.3) & 95.1 & 74.3\\
SlowFast-4$\times$16 \cite{Feichtenhofer2019SlowFastNF} & 36.1G$\times$30 & \textbf{75.6} (92.1) & - & -\\
\midrule
ARTNet \cite{wang2018appearance} & 23.5$\times$250 & 69.2 (88.3) & 94.3 & 70.9\\
Zhao \emph{et. al.} \cite{zhao2018recognize} & - & 71.5 (89.9) & 95.9 & -\\
TSN$\rm _{RGB}$ \cite{wang2016temporal} & 53G$\times$10 & 69.1 (88.7) & 91.1 & -\\
OFF$\rm _{RGB}$ \cite{sun2018optical} & - & - & 93.3 & -\\
TSM \cite{lin2019tsm} & 43G$\times$30 & 74.1 (91.2) & 95.9 & 73.5\\
TEA \cite{Li_2020_CVPR} & 35G$\times$30 & 75.0 (91.8) & 96.9 & 73.3\\
\midrule
{\color{mygray}I3D$\rm _{RGB+Flow}$ \cite{carreira2017quo}} & {\color{mygray}-} & {\color{mygray}74.2 (91.3)} & {\color{mygray}\textbf{98.0}} & {\color{mygray}\textbf{80.7}}\\
{\color{mygray}TSN$\rm _{RGB+Flow}$ \cite{wang2016temporal}} & {\color{mygray}-} & {\color{mygray}73.9 (91.1)} & {\color{mygray}97.0} & {\color{mygray}-}\\
{\color{mygray}OFF$\rm _{RGB+Flow}$ \cite{sun2018optical}} & {\color{mygray}-} & {\color{mygray}-} & {\color{mygray}96.0} & {\color{mygray}74.2}\\
\midrule
PAN$\rm _{Lite}$ (Ours) & 47G$\times$2 & 73.1 (91.1) & 96.0 & 74.5\\
PAN$\rm _{Full}$ (Ours) & 88G$\times$2 & 74.4 (91.6) & 96.5 & 77.0\\
PAN$\rm _{En}$ (Ours) & 135G$\times$2 & \underline{75.3} (\textbf{92.4}) & \underline{97.2} & \underline{77.3}\\
\bottomrule
\end{tabular}
\begin{tablenotes}
\footnotesize
\item * For fair comparison, all the results on UCF101 and HMDB51 are obtained with Kinetics400 pre-train.
\end{tablenotes}
\end{threeparttable}
\end{adjustbox}
\end{center}
\end{table}

\subsection{Visualization and Discussion} \label{section:visualization}

In this section, we present the visualization results of two adjacent frames and their corresponding PA and optical flow in Fig.~\ref{fig:vis}. PA can model motion information as optical flow does, but is more advanced for action recognition task with its special characteristic. Similar to optical flow, the computed PA highlights the moving objects and suppresses the stationary background, which visually demonstrates that our PA well characterizes the instantaneous motion. Differently, our PA focuses more on the motion boundaries instead of the whole moving areas. This aligns with our motivation that modeling small displacements of motion boundaries matters most for action recognition task. For more visualization results, please refer to our supplementary material.


Visually, our PA contains more noise than optical flow. We speculate that it is because the regularization term, penalizing high variations to obtain smooth displacement fields, is not applied for efficiency concerns, while it is used in conventional optical flow method \cite{zach2007duality}. Although our PA has been proved superior to several optical flow methods in terms of motion modeling efficiency and action recognition accuracy, we have not fully exploited its potential because of the noise interference. So in the future, we plan to design the noise mitigation method to obtain more smooth PA.




\section{Conclusion}

In this paper, we shed light on fast action recognition by lifting the reliance on optical flow. We design a concise motion cue called \emph{Persistence of Appearance} (PA) to capture motion information directly from RGB frames. In contrast to optical flow, our PA is more effective by focusing more on modeling the small displacements of motion boundaries, and it is more efficient by simply calculating the pixel-wise differences between two adjacent frames in feature space. Its efficiency, effectiveness and flexibility have been well elaborated by extensive theoretical support (Sec.~\ref{section:sub_PA}), experimental support (Sec.~\ref{section:PA_ablation}) and visualization support (Sec.~\ref{section:visualization}). The motion modeling speed of our PA reaches 1000$\times$ faster than that of conventional optical flow method (8196fps vs 8fps). To further aggregate the short-term dynamics in PA to long-term dynamics, we also propose a temporal fusion strategy named \emph{Various-timescale Aggregation Pooling} (VAP), which enables the network to capture long-range various-timescale interdependencies. The proposed PA and VAP are finally incorporated to form a unified framework call \emph{Persistent Appearance Network} (PAN). Extensive experiments on six challenging benchmarks demonstrate that our proposed PAN achieves the state-of-the-art recognition performance. Most importantly, it significantly accelerates the inference process of action recognition with the powerful motion cue PA.

\ifCLASSOPTIONcaptionsoff
  \newpage
\fi



\bibliographystyle{IEEEtran}

\clearpage



\section*{Supplementary material}

# −Supplementary Material−
## PAN: Towards Fast Action Recognition via Learning Persistence of Appearance

Can Zhang, Yuexian Zou*, *Senior Member, IEEE,* Guang Chen, and Lei Gan

## I. IMPLEMENTATION DETAILS OF TABLE I AND TABLE II

In this section, we describe the implementation details for the comparative experiments of two encoding schemes (see Table I) and the comparative experiments of PA with other mainstream optical flow computation methods (see Table II). To have an apple-to-apple comparison, the implementation details of all the experiments are identical. We set $N = 8$ and $m = 6$, *i.e.*, 8 sampled "6-frame stack" RGB frames are sampled as input.

To measure the efficiency of the motion modeling methods, we use three evaluation metrics including the computational cost (FLOPs), the number of parameters (#Param) and inference speed (Speed). All the efficiency metrics in these two tables are measured on a single NVIDIA TITAN X GPU with 1 mini batch size and 1 CPU thread. As the I/O is most relevant to hardware and operating system, the runtime speeds are reported without considering I/O.

To evaluate the performance of the motion representation (encoded PA or optical flow) on action recognition task, we obtain accuracy results on UCF101 split1 using the motion representation as the input modality. Following the TSN manner, we first sample frames from evenly divided video segments, then these frames are fed into the motion modeling module and backbone CNN (ResNet-50) sequentially. Finally, the output activations are averaged as the final prediction scores. Note that all the accuracy results in both tables are measured with the same network settings: initial learning rate: 0.001; total epochs: 80 (decreases lr by 10 after every 30 epochs); mini batchsize: 16; dropout ratio: 0.7; pretrain: ImageNet.

## II. VISUALIZATION RESULTS OF PA

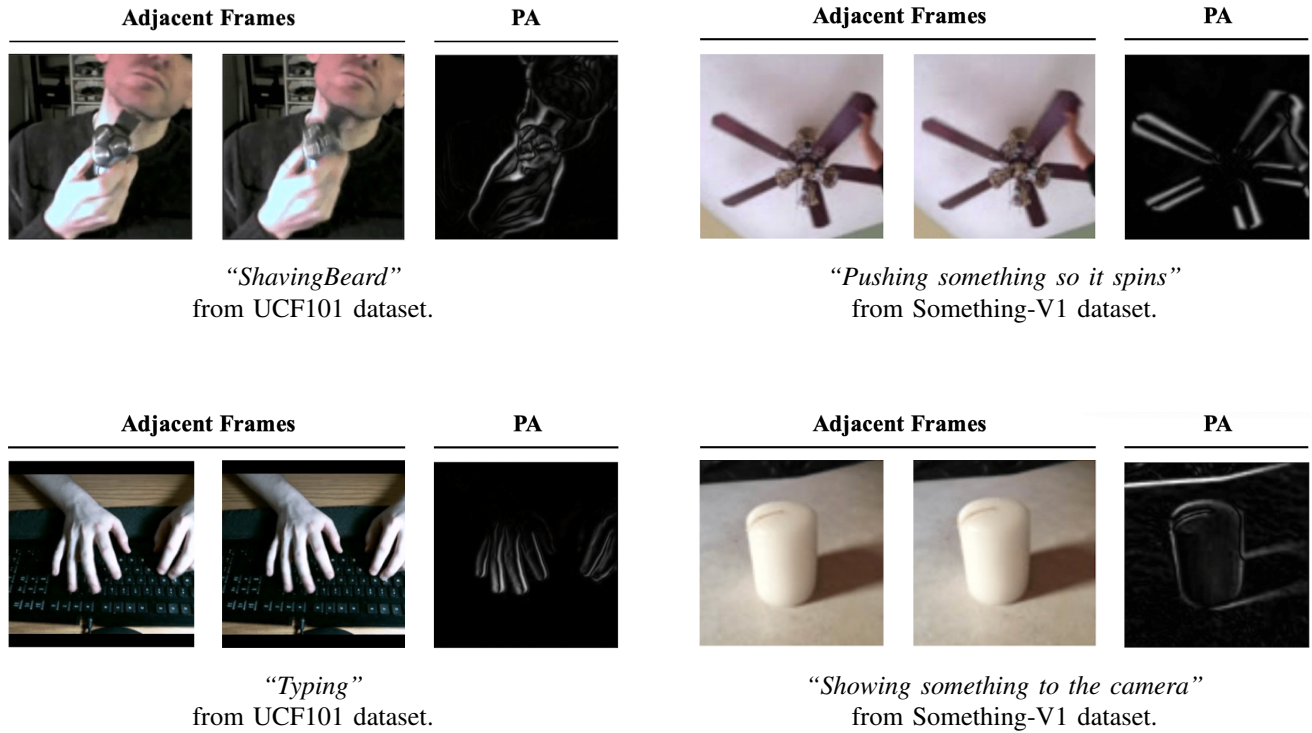

"ShavingBeard" from UCF101 dataset.

"Pushing something so it spins" from Something-V1 dataset.

"Typing" from UCF101 dataset.

"Showing something to the camera" from Something-V1 dataset.

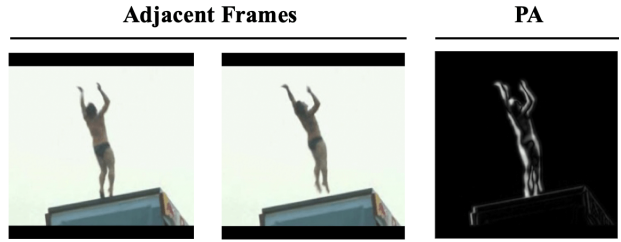

*"CliffDiving"*
from UCF101 dataset.

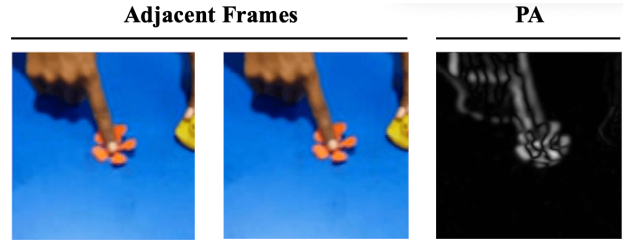

*"Moving something closer to something"*
from Something-V1 dataset.

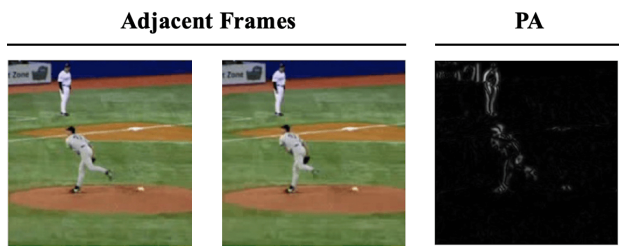

*"BaseballPitch"*
from UCF101 dataset.

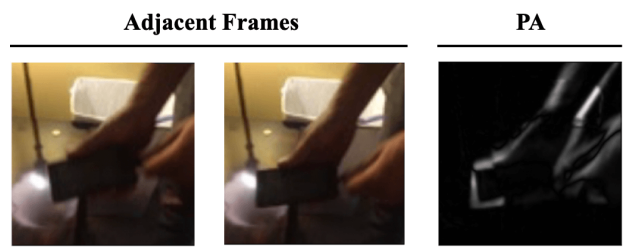

*"Plugging something into something"*
from Something-V1 dataset.

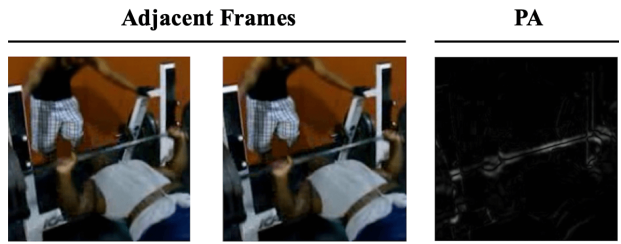

*"BenchPress"*
from UCF101 dataset.

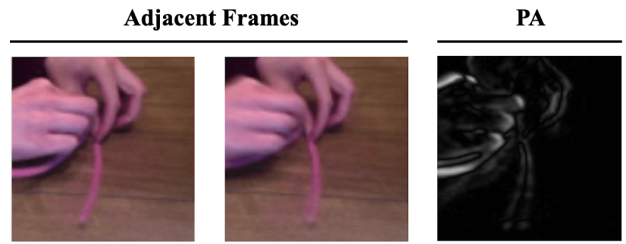

*"Bending something so that it deforms"*
from Something-V1 dataset.

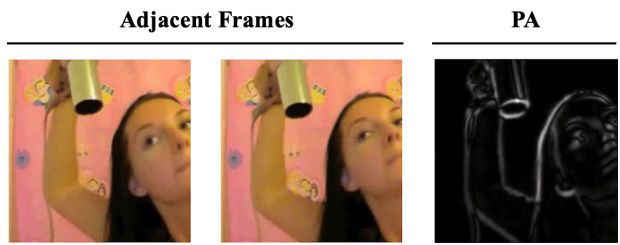

*"BlowDryHair"*
from UCF101 dataset.

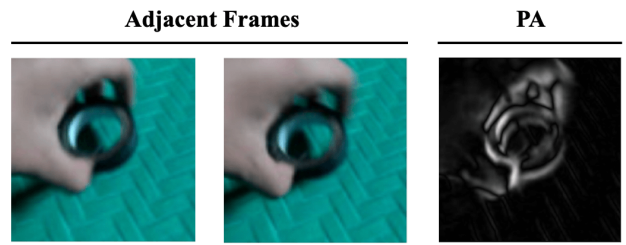

*"Spinning something that quickly stops spinning"*
from Something-V1 dataset.

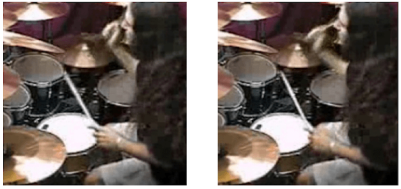 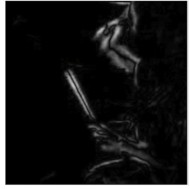 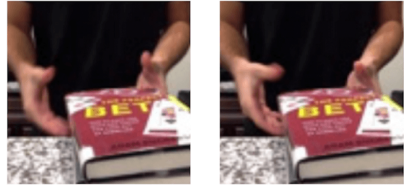 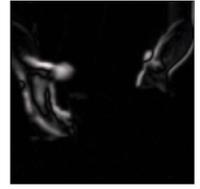

*"Drumming"* from UCF101 dataset.

*"Lifting something with something on it"* from Something-V1 dataset.

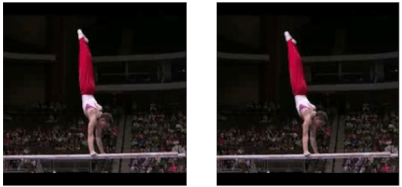 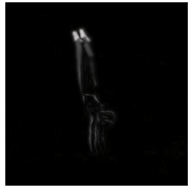 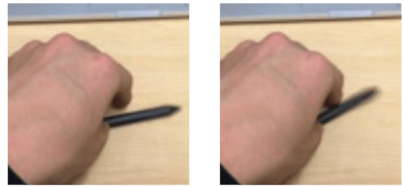 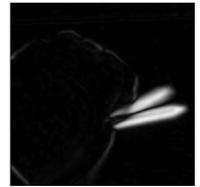

*"ParallelBars"* from UCF101 dataset.

*"Spinning something that quickly stops spinning"* from Something-V1 dataset.

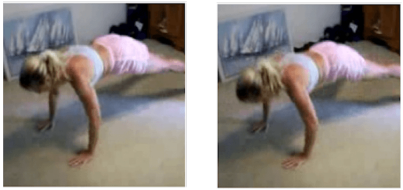 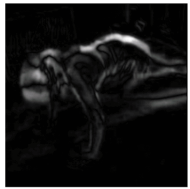 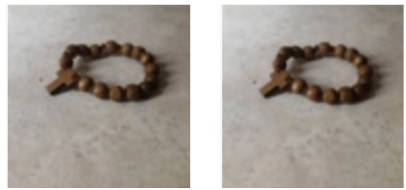 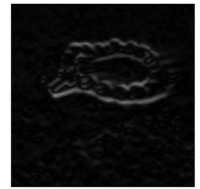

*"PushUps"* from UCF101 dataset.

*"Turning the camera upwards while filming something"* from Something-V1 dataset.

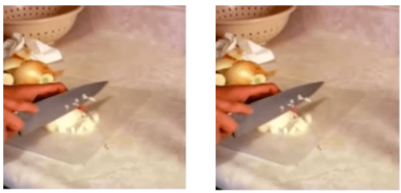 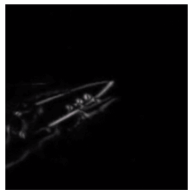 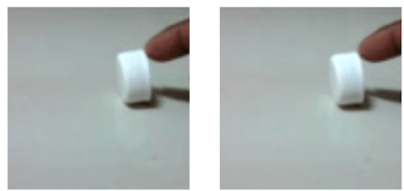 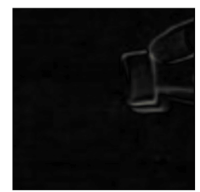

*"CuttingInKitchen"* from UCF101 dataset.

*"Tipping something over"* from Something-V1 dataset.

4

## III. LEARNED TIMESCALE-WISE WEIGHTS $w$ OF VAP

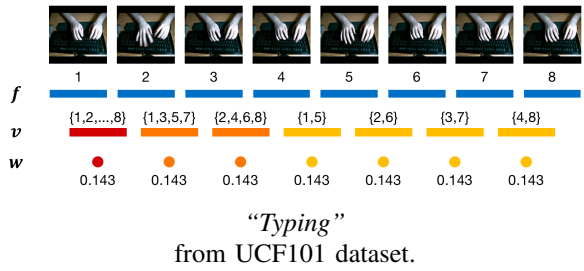

*"Typing"*
from UCF101 dataset.

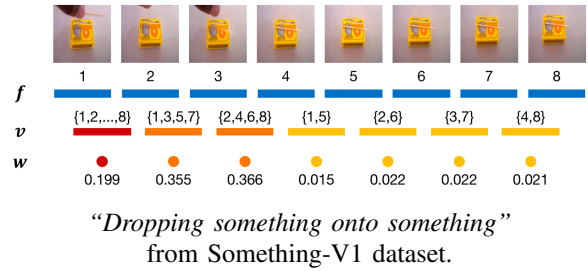

*"Dropping something onto something"*
from Something-V1 dataset.

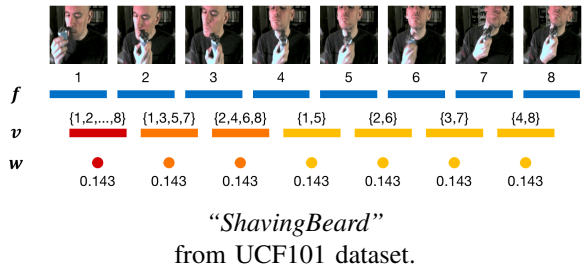

*"ShavingBeard"*
from UCF101 dataset.

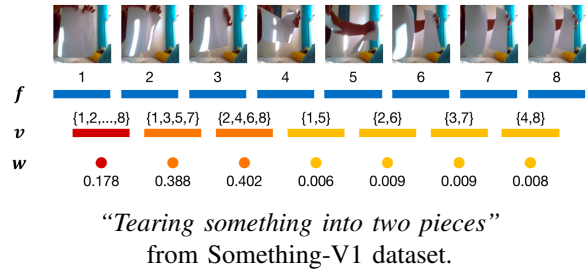

*"Tearing something into two pieces"*
from Something-V1 dataset.

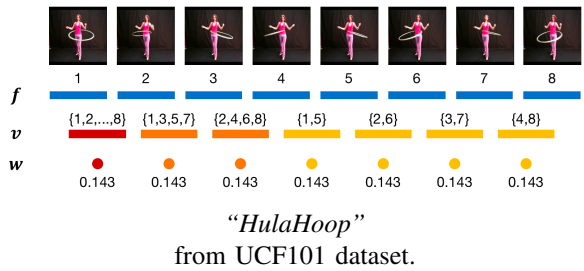

*"HulaHoop"*
from UCF101 dataset.

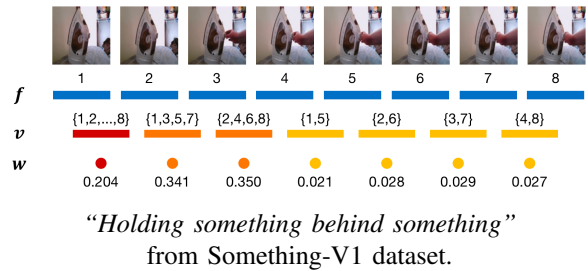

*"Holding something behind something"*
from Something-V1 dataset.

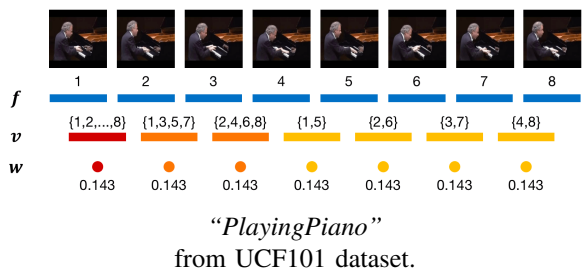

*"PlayingPiano"*
from UCF101 dataset.

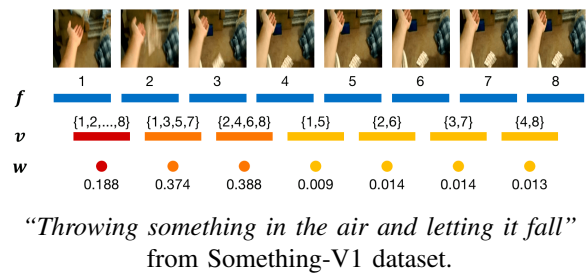

*"Throwing something in the air and letting it fall"*
from Something-V1 dataset.

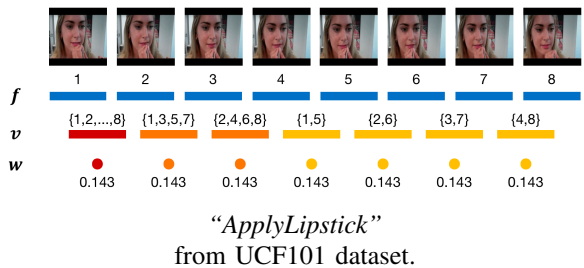

*"ApplyLipstick"*
from UCF101 dataset.

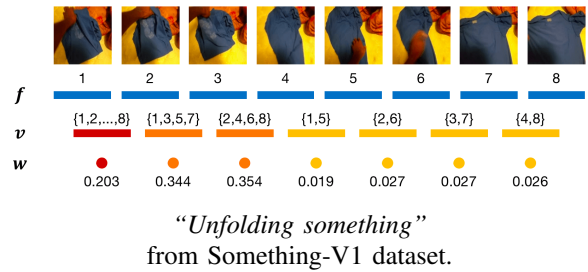

*"Unfolding something"*
from Something-V1 dataset.



\end{document}